\documentclass{article}

\usepackage{cite}
\usepackage{amsmath,amssymb,amsfonts}
\usepackage{algorithmic}
\usepackage{graphicx}
\usepackage{textcomp}
\usepackage{xcolor}
\usepackage[hyphens]{url}
\usepackage{multirow}
\usepackage{booktabs}
\usepackage{pifont}
\usepackage{array}
\usepackage{xspace}
\usepackage{soul}
\usepackage{subcaption}
\usepackage{fancyhdr}
\usepackage{colortbl}
\usepackage{makecell}
\usepackage[bookmarks=true,breaklinks=true,letterpaper=true,colorlinks,citecolor=blue,linkcolor=blue,urlcolor=blue]{hyperref}
\usepackage{listings}

\usepackage{microtype}

\usepackage[accepted]{mlsys2024}

\usepackage{fancyhdr} 
\newcommand{\sect}[1]{Section~\ref{#1}}

\newcommand{\eqn}[1]{Equation~\ref{#1}}

\newcommand{\fig}[1]{Figure~\ref{#1}}

\newcommand{\tab}[1]{Table~\ref{#1}}

\makeatletter
\DeclareRobustCommand\onedot{\futurelet\@let@token\@onedot}
\def\@onedot{\ifx\@let@token.\else.\null\fi\xspace}

\def\eg{\emph{e.g}\onedot} 
\def\ie{\emph{i.e}\onedot} 
 
 \def\vs{\emph{vs}\onedot}

\makeatother

\definecolor{mydarkblue}{rgb}{0,0.08,1}
\definecolor{mydarkgreen}{rgb}{0.02,0.6,0.02}
\definecolor{mydarkred}{rgb}{0.8,0.02,0.02}
\definecolor{mydarkorange}{rgb}{0.40,0.2,0.02}
\definecolor{mypurple}{rgb}{111,0,255}
\definecolor{myred}{rgb}{1.0,0.0,0.0}
\definecolor{mygold}{rgb}{0.75,0.6,0.12}
\definecolor{myblue}{rgb}{0,0.2,0.8}
\definecolor{mylightblue}{rgb}{0.827,0.937,0.945}
\definecolor{mydarkgray}{rgb}{0.66,0.66,0.66}
\definecolor{mygray}{rgb}{0.85,0.85,0.85}

\def\x{$\times$\xspace}

\def\contextstage{prefilling stage\xspace}
\def\decodingstage{decoding stage\xspace}
\def\system{QServe\xspace}
\def\smoothattention{SmoothAttention\xspace}
\def\algo{QoQ\xspace}

\definecolor{LightGray}{gray}{0.98}
\definecolor{codegreen}{rgb}{0,0.6,0}
\definecolor{codegray}{rgb}{0.5,0.5,0.5}
\definecolor{codepurple}{rgb}{0.58,0,0.82}
\definecolor{backcolour}{rgb}{0.95,0.95,0.92}

\lstdefinestyle{mystyle}{
    frame=single,                     
    framesep=1mm,                     
    framerule=0.4pt,                  
    rulecolor=\color{black!80},       
    xleftmargin=5pt,                  
    xrightmargin=5pt,                     backgroundcolor=\color{backcolour},
    tabsize=2,                        
    breaklines=true,                  
    breakatwhitespace=false,          
    keepspaces=true,                  
    basicstyle=\ttfamily\footnotesize,
    numbers=left,                     
    numbersep=5pt,                    
    numberstyle=\footnotesize\color{codegray}, 
    commentstyle=\color{codegreen},   
    keywordstyle=\color{magenta},         stringstyle=\color{codepurple},   
    identifierstyle=\color{black},    
    showspaces=false,                 
    showstringspaces=false,           
    showtabs=false,                   
    captionpos=b                      
}

\mlsystitlerunning{QServe: \textbf{\texttt{W4A8KV4}} Quantization and System Co-design for Efficient LLM Serving}

\usepackage{graphicx}
\usepackage{hyperref}
\usepackage{eso-pic}
\usepackage{xparse}

\newlength{\badgewidth}
\newlength{\badgegap}
\newcommand{\badgeList}{}

\NewDocumentCommand{\addTopRightBadge}{O{} m}{%
\gappto{\badgeList}{\href{#1}{\includegraphics[width=\badgewidth]{#2}}\hspace{\badgegap}}%
}

\newcommand{\placeTopRightBadges}{%
\AddToShipoutPictureBG*{%
\put(\LenToUnit{\paperwidth - 1.5cm - \badgewidth},\LenToUnit{\paperheight - 2cm}){%
\makebox[0pt][r]{\badgeList}%
}%
}%
}

\addTopRightBadge{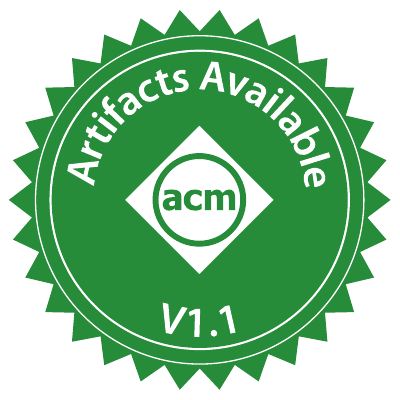}
\addTopRightBadge{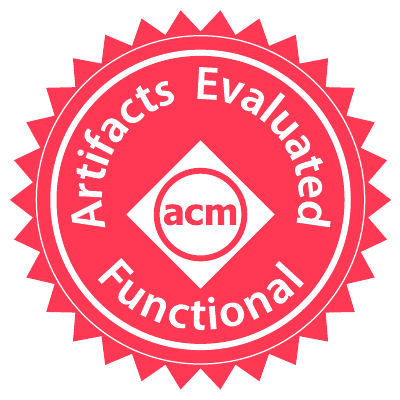}

\placeTopRightBadges

\begin{document}

\twocolumn[
\mlsystitle{QServe: \textbf{\texttt{W4A8KV4}} Quantization and System Co-design for Efficient LLM Serving}

\mlsyssetsymbol{equal}{*}

\begin{mlsysauthorlist}
\mlsysauthor{Yujun Lin}{equal,mit}
\mlsysauthor{Haotian Tang}{equal,mit}
\mlsysauthor{Shang Yang}{equal,mit}
\mlsysauthor{Zhekai Zhang}{mit}
\mlsysauthor{Guangxuan Xiao}{mit}
\mlsysauthor{Chuang Gan}{umass,ibm}
\mlsysauthor{Song Han}{mit,nv}
\end{mlsysauthorlist}

\mlsysaffiliation{mit}{MIT}
\mlsysaffiliation{umass}{UMass Amherst}
\mlsysaffiliation{ibm}{MIT-IBM Watson AI Lab}
\mlsysaffiliation{nv}{NVIDIA}

\begin{center}
\url{https://hanlab.mit.edu/projects/qserve} 
\end{center}

\mlsyscorrespondingauthor{Yujun Lin}{yujunlin@mit.edu}
\mlsyscorrespondingauthor{Haotian Tang}{kentang@mit.edu}
\mlsyscorrespondingauthor{Shang Yang}{shangy@mit.edu}
\mlsyscorrespondingauthor{Song Han}{songhan@mit.edu}

\mlsyskeywords{Large Language Model, LLM, Quantization}

\vskip 0.3in

\begin{abstract}

Quantization can accelerate large language model (LLM) inference. Going beyond \texttt{INT8} quantization, the research community is actively exploring even lower precision, such as \texttt{INT4}. Nonetheless, state-of-the-art \texttt{INT4} quantization techniques only accelerate low-batch, edge LLM inference, failing to deliver performance gains in large-batch, cloud-based LLM serving. We uncover a critical issue: existing \texttt{INT4} quantization methods suffer from significant runtime overhead (20-90\%) when dequantizing either weights or partial sums on GPUs. To address this challenge, we introduce QoQ, a \texttt{W4A8KV4} quantization algorithm with 4-bit weight, 8-bit activation, and 4-bit KV cache. \algo stands for \textit{quattuor-octō-quattuor}, which represents 4-8-4 in Latin. QoQ is implemented by the \system inference library that achieves measured speedup. The key insight driving  \system is that the efficiency of LLM serving on GPUs is critically influenced by operations on low-throughput CUDA cores. Building upon this insight, in \algo algorithm, we introduce \textit{progressive quantization} that can allow low dequantization overhead in \texttt{W4A8} GEMM. Additionally, we develop \textit{SmoothAttention} to effectively mitigate the accuracy degradation incurred by 4-bit KV quantization. In the \system system, we perform \textit{compute-aware weight reordering} and take advantage of \textit{register-level parallelism} to reduce dequantization latency. We also transfer theoretical memory saving brought by KV4 attention into measured speedup using \system. 
As a result, \system improves the maximum achievable serving throughput of Llama-3-8B by \textbf{1.2\x} on A100, \textbf{1.4\x} on L40S; and Qwen1.5-72B by \textbf{2.4\x} on A100, \textbf{3.5\x} on L40S, compared to TensorRT-LLM. Remarkably, \system on L40S GPU can achieve even higher throughput than TensorRT-LLM on A100. Code is released at \url{https://github.com/mit-han-lab/omniserve}. %

\end{abstract}

]

\printAffiliationsAndNotice{\mlsysEqualContribution} %

\section{Introduction}
\label{sect:intro}

\begin{figure}[t]
    \centering
    \includegraphics[width=\linewidth]{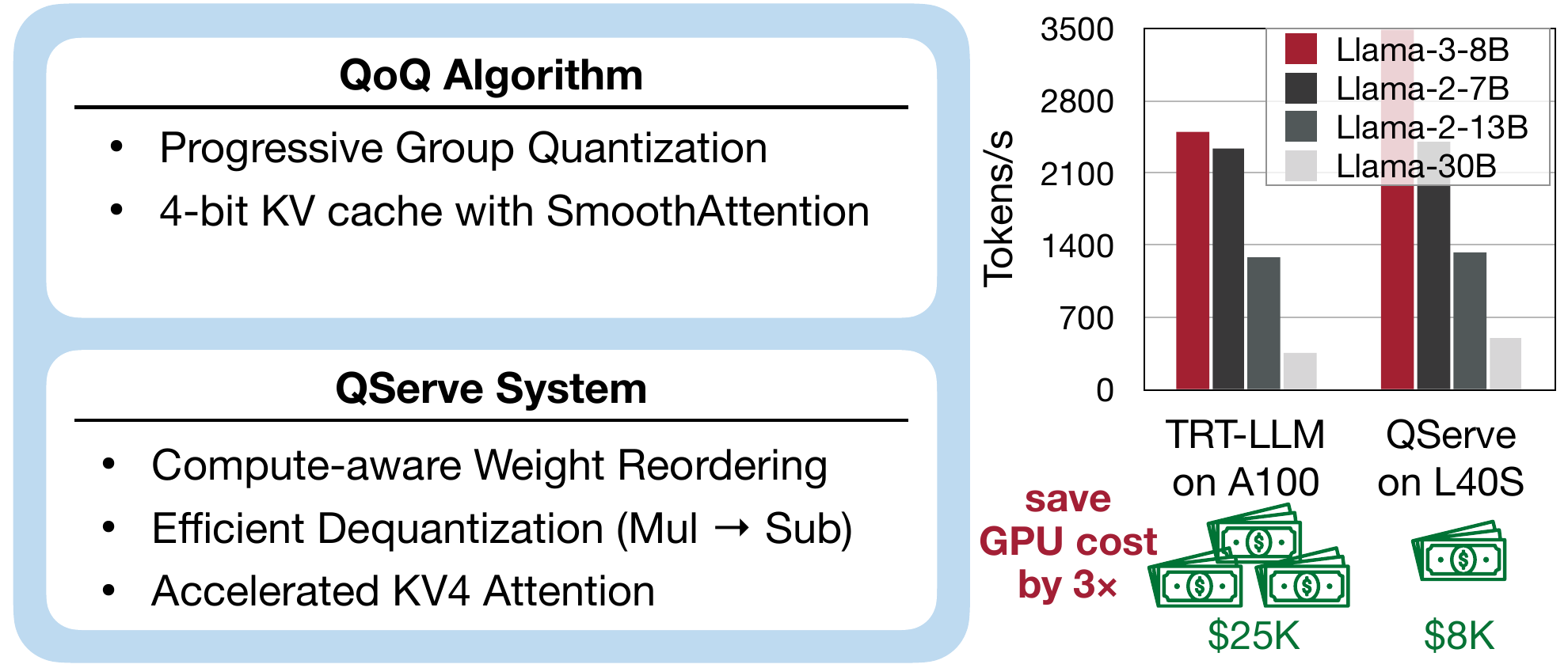}
    \vspace{-18pt}
    \caption{\system achieves higher throughput when running Llama models on L40S compared with TensorRT-LLM on A100, effectively saves the dollar cost for LLM serving by \textbf{3\x} through system-algorithm codesign. See Table~\ref{tab:abs_main_speed} for absolute throughput numbers and precision choices in TensorRT-LLM.} %
    \label{fig:intro:teaser}
    \vspace{-18pt}
\end{figure}

Large language models (LLMs) have demonstrated remarkable capability across a broad spectrum of tasks, exerting a profound influence on our daily lives. 
However, the colossal size of LLMs makes their deployment extremely challenging, necessitating the adoption of quantization techniques for efficient inference. State-of-the-art integer quantization algorithms can be divided into three categories: 8-bit weight and 8-bit activation (\textbf{\texttt{W8A8}}), 4-bit weight and 16-bit activation (\textbf{\texttt{W4A16}}), 4-bit weight 4-bit activation (\textbf{\texttt{W4A4}}) quantization. The former two methods are considered nearly lossless in terms of accuracy. In contrast, \textbf{\texttt{W4A4}} quantization introduces a notable accuracy degradation, although it is anticipated to offer superior throughput in return by mapping its computations onto high-throughput 4-bit tensor cores. Unfortunately, this anticipated performance boost has not been consistently observed across current GPU platforms. %
For instance, the state-of-the-art \textbf{\texttt{W4A4}} serving system, Atom~\cite{zhao2023atom}, exhibits 20-25\% lower performance than its \textbf{\texttt{W4A16}} and \textbf{\texttt{W8A8}} counterpart in TensorRT-LLM when running the Llama-2-7B~\cite{touvron2023llama2} model on A100 GPUs. That said, the research community has yet to find a precision combination superior to \textbf{\texttt{W4A16}} and \textbf{\texttt{W8A8}} for efficient cloud LLM serving.

In this paper, we reveal a critical observation: current 4-bit integer quantization methods experience significant overhead, ranging from 20\% to 90\%, during the dequantization of weights or partial sums %
on current-generation GPUs. For example, \textbf{\texttt{W4A16}} quantization performs computation on \texttt{FP16} tensor cores while the weights are in \texttt{INT4}, so weight dequantization is required in the GEMM kernel. On the other hand, for \textbf{\texttt{W4A4}} quantization, to achieve reasonable accuracy, \textbf{\texttt{W4A4}} methods must apply \textbf{per-group} quantization to both weights and activation, sharing \texttt{FP16} scaling factors on a sub-channel basis. For example, the state-of-the-art \textbf{\texttt{W4A4}} quantization method, QuaRot~\cite{ashkboos2024quarot}, reports a significant 0.2 perplexity degradation after switching from per-group quantization to per-channel quantization. This per-group quantization design requires an integer to floating-point dequantization for partial sums (since \texttt{INT4} tensor cores produce \texttt{INT32} partial sums), which operates on the slower CUDA cores within the sequential main loop of \textbf{\texttt{W4A4}} GEMM. On data center GPUs like A100, a CUDA core operation is as expensive as \textbf{50} \texttt{INT4} tensor core operations. Therefore, reducing bit precision not necessarily speeds up LLM inference. 

To achieve optimal LLM serving throughput, We introduce \algo (Quattuor-Octō-Quattuor, or 4-8-4 in Latin) algorithm which quantizes LLMs to \textbf{\texttt{W4A8KV4}} precision: 4-bit weights, 8-bit activations and 4-bit KV caches. Additionally, we present \system, which provides efficient system support for \textbf{\texttt{W4A8KV4}} quantization. 

In the \algo algorithm, we introduce \textit{progressive group quantization}. This method first quantizes weights to 8 bits using per-channel \texttt{FP16} scales with a \textit{protective range} of $[-119, 119]$, then quantizes these 8-bit intermediates to 4 bits. This approach ensures that all GEMMs are performed on \texttt{INT8} tensor cores. Additionally, we mitigate accuracy loss from \texttt{KV4} quantization through \textit{SmoothAttention}, which shifts the challenge of activation quantization from keys to queries, the latter of which are not quantized. 

In the \system system, the \textit{protective range} in progressive group quantization enables full \textit{register-level parallelism} during \texttt{INT4} to \texttt{INT8} dequantization, using a \textit{subtraction after multiplication} computation order. Furthermore, we propose \textit{compute-aware weight reordering} to minimize pointer arithmetic overhead on CUDA cores during W4A8 GEMM operations. Additionally, we delay the turning point of the CUDA core roofline and decrease the computational intensity of \texttt{KV4} attention at the same time. This ensures that the attention operator remains within the memory-bound region, where low-bit quantization can effectively enhance throughput.

We evaluate seven widely-used LLMs using \system on A100 and L40S GPUs, and compare their maximum achievable throughput against state-of-the-art systems, including TensorRT-LLM (in \texttt{FP16}, \textbf{\texttt{W8A8}}, and \textbf{\texttt{W4A16}} configurations), Atom~\cite{zhao2023atom} (in \textbf{\texttt{W4A4}}), and QuaRot~\cite{ashkboos2024quarot} (in \textbf{\texttt{W4A4}}). On A100 GPUs, \system achieves \textbf{1.2-2.4\x} higher throughput over the best-performing configuration of TensorRT-LLM, and \textbf{2.5-2.9\x} higher throughput compared to Atom and QuaRot. On L40S GPUs, \system records an even more significant \textbf{1.5-3.5\x} throughput improvement over TensorRT-LLM. Notably, we manage to accommodate the same batch size on the L40S while consistently achieving higher serving throughput than TensorRT-LLM on A100 which is 3$\times$ more expensive for six of the eight models tested.

\section{Background}
\subsection{Large Language Models}
\label{sect:background:preliminaries}

Large Language Models (LLMs) are a family of causal transformer models with multiple identically-structured layers. Each layer combines an attention block, a feed-forward network (FFN) and normalization layers. The input of each layer, $\mathbf{x}$, is an $N\times HD$ tensor, where $N$ is the number of input tokens, $H$ represents the number of attention heads, and $D$ is the hidden dimension for each head. Serving LLMs involves two stages: the \contextstage, where all prompt tokens are presented simultaneously ($N > 1$ for each request), and the \decodingstage, where the model only processes one token at a time for each prompt ($N = 1$ for each request).

In attention blocks, $\mathbf{x}$ first undergoes linear projection to obtain $\mathbf{q}\in\mathbb{R}^{N\times HD},\mathbf{k},\mathbf{v}\in\mathbb{R}^{N\times H_{KV}D}$, where $H_{KV}$ is the number of key/value heads. We have $H = H_{KV}$ in the standard multi-head attention (MHA), while recent methods~\cite{touvron2023llama2,jiang2023mistral,jiang2024mixtral} also employ grouped-query attention (GQA)~\cite{ainslie2023gqa} with $H = rH_{KV} (r\in \mathbb{Z})$. We concatenate $\mathbf{k},\mathbf{v}$ with pre-computed \textit{KV cache} features of $S$ previous tokens to obtain $\mathbf{K}, \mathbf{V}\in\mathbb{R}^{(S+N)\times H_{KV}D}$ and compute attention using:

\vspace{-10pt}
\begin{equation}
\small
\mathbf{o}_{h} = \text{softmax}\left(\frac{\mathbf{q}_{h}\mathbf{K}_{h_{KV}}^T}{\sqrt{D}}\right)\mathbf{V}_{h_{KV}},\quad h_{KV}=\left\lfloor\frac{h}{r}\right\rfloor
\label{eqn:background:attn}
\end{equation}
\vspace{-10pt}

The result $\mathbf{o}$ is multiplied with an output projection matrix $\mathbf{W}_O \in \mathbb{R}^{HD \times HD}$, and the product is added to $\mathbf{x}$ as the input of FFN. The FFN is composed of linear projection and activation layers and it does not mix features between tokens. 

\subsection{Integer Quantization}

Integer quantization maps high-precision numbers to discrete levels. The process can be formulated as:

\begin{equation}
\resizebox{0.9\linewidth}{!}{
$\displaystyle 
\small
    \mathbf{Q}_{\mathbf{X}} = \left\lceil \frac{\mathbf{X}}{s}+z\right\rfloor, s=\frac{\mathbf{X}_{\max}-\mathbf{X}_{\min}}{q_{\max}-q_{\min}}, z= \left\lceil q_{\min}-\frac{\mathbf{X}_{\min}}{s}\right\rfloor
$%
}
\label{eqn:background:quantization}
\end{equation}
\vspace{-15pt}

where $\mathbf{X}$ is the floating point tensor, $\mathbf{Q}_{\mathbf{X}}$ is its $n$-bit quantized counterpart, $s$ is the scaling factor and $z$ is the zero point. Thus, the dequantized tensor can be represented as,

\vspace{-10pt}
\begin{equation}
\small
\hat{\mathbf{X}} = Q\left(\mathbf{X}\right) = \left(\mathbf{Q}_{\mathbf{X}} - z\right) \cdot s
\label{eqn:background:dequantization}
\end{equation}
\vspace{-18pt}

This is known as  \textit{asymmetric} quantization, where $\mathbf{X}_{\max} = \max\left(\mathbf{X}\right), \mathbf{X}_{\min} = \min\left(\mathbf{X}\right)$, and $q_{\max} - q_{\min} = 2^{n}-1$ for integer quantization. Equation \ref{eqn:background:quantization} can be further simplied to \textit{symmetric} quantization, where $z = \mathbf{0}$, $\mathbf{X}_{\max} = -\mathbf{X}_{\min} = \max\left|\mathbf{X}\right| $, and $q_{\max} - q_{\min} = 2^{n} -2$ .

In this paper, we denote $x$-bit weight, $y$-bit activation and $z$-bit KV cache quantization in LLMs as \textbf{\texttt{WxAyKVz}}, and use the abbreviated notation \textbf{\texttt{WxAy}} if \texttt{y=z}.  Apart from bit precision, quantization can also be applied at various granularities. \textit{Per-tensor} quantization shares $s$ and $z$ across the entire tensor. \textit{Per-channel} quantization for weights or \textit{per-token} quantization for activations means that $s$ and $z$ are shared within each \textit{row} of tensor. \textit{Per-group} quantization further reduces the degree of parameter sharing by using different $s$ and $z$ for every $g$ columns within each row, where $g$ is the group size.

\section{Motivation}
\label{sect:motivation}

In this paper, we denote $x$-bit weight, $y$-bit activation and $z$-bit KV cache quantization in LLMs as \textbf{\texttt{WxAyKVz}}, and use the abbreviated notation \textbf{\texttt{WxAy}} if \texttt{y=z}.  Apart from bit precision, quantization can also be applied at various granularities. \textit{Per-tensor} quantization shares $s$ and $z$ across the entire tensor. \textit{Per-channel} quantization for weights or \textit{per-token} quantization for activations means that $s$ and $z$ are shared within each \textit{row} of tensor. \textit{Per-group} quantization further reduces the degree of parameter sharing by using different $s$ and $z$ for every $g$ columns within each row, where $g$ is the group size.

Weight and KV cache quantization (\textit{e.g.} \textbf{\texttt{W4}},\textbf{ \texttt{KV4}}) can reduce the memory footprint in LLM serving. Quantizing both weight and activation (\textit{e.g.} \textbf{\texttt{W8A8}}) can also improve the peak computation throughput. Choosing the right precision for LLM deployment is a difficult task. Existing solutions can be divided into three categories: \textbf{\texttt{W4A16}} (per-group), \textbf{\texttt{W8A8}} (per-channel weight + per-token activation), \textbf{\texttt{W4A4}} (per-group). We will demonstrate in this section why \textbf{\texttt{W4A8KV4}} is a superior choice.

\subsection{\textbf{\texttt{W4A8KV4}} Has Superior Roofline Over \textbf{\texttt{W8A8}}, \textbf{\texttt{W4A16}}}

\begin{figure}
    \centering
    \includegraphics[width=\linewidth]{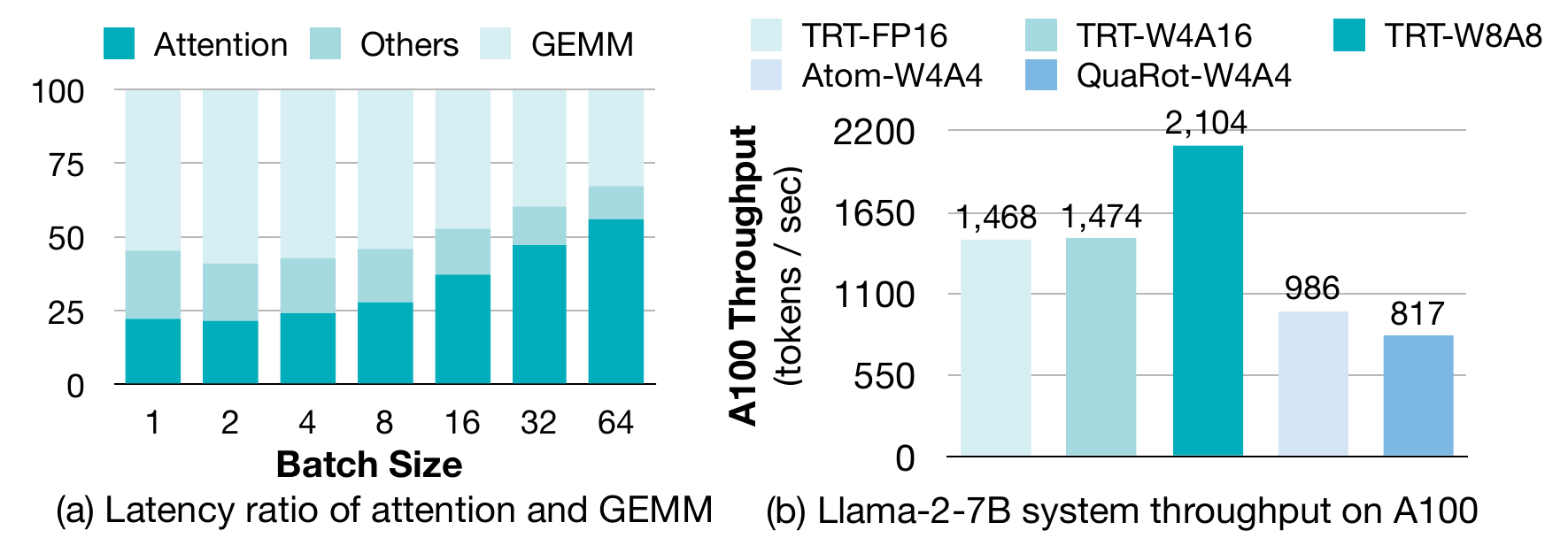}
    \vspace{-20pt}
    \caption{\textbf{Left}: Both attention and GEMM are crucial for end-to-end LLM latency. \textbf{Right}: Despite 2$\times$ higher theoretical peak performance, \textbf{\texttt{W4A4}} systems significantly lag behind TRT-LLM-\textbf{\texttt{W8A8}} in efficiency.} %
    \label{fig:motivation:precision_choice}
    \vspace{-10pt}
\end{figure}

\begin{figure}[t]
    \centering
    \includegraphics[width=\linewidth]{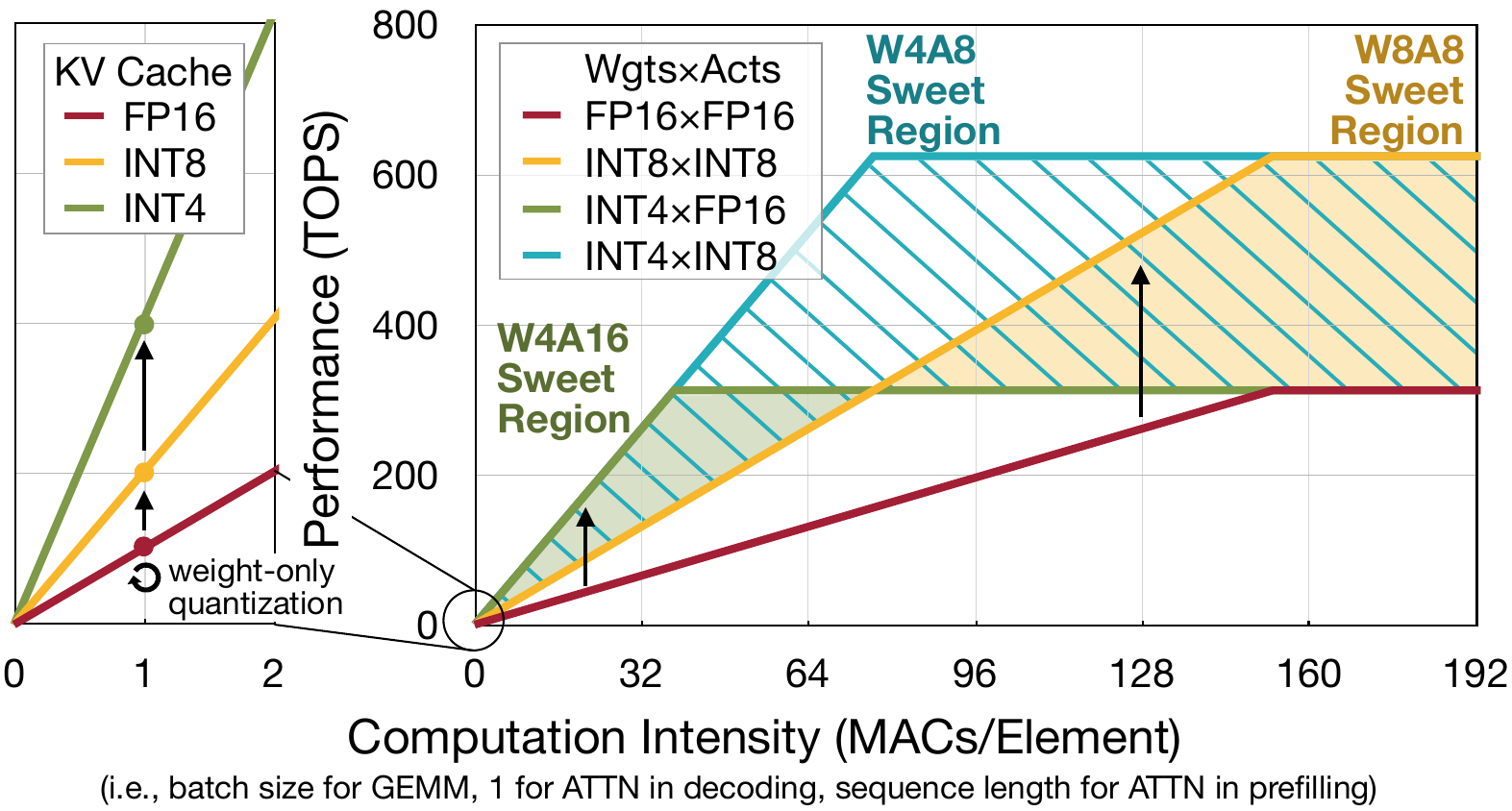}
    \vspace{-20pt}
    \caption{\textbf{A100 roofline for LLM serving}: for GEMM layers, the \textbf{\texttt{W4A8}} roofline dominates both \textbf{\texttt{W4A16}} and \textbf{\texttt{W8A8}} across different batch sizes; for attention layers, 4-bit quantization improves theoretical peak performance.}
    \label{fig:motivation:roofline}
    \vspace{-20pt}
\end{figure}

We begin our exploration through roofline analysis. As in Figure \ref{fig:motivation:precision_choice}a, when considering real-world conversations with 1024 input tokens and 512 output tokens, attention and GEMM account for most of the runtime when deploying LLMs. Furthermore, the runtime of the \decodingstage is approximately 6\x that of the \contextstage. Therefore, we focus our analysis on the attention and GEMM within the \decodingstage. 

For an $m\times n\times k$ GEMM problem, the computation intensity (defined as MACs/element) is approximately $m$ when $n, k$ are much larger than $m$. This situation applies to LLM decoding stage, since $m$ is number of sequences and $n, k$ are channel sizes. According to the A100 roofline\footnote{A100 has a peak FP16/INT8/INT4 tensor core performance of 312/624/1248 TOPS and a DRAM bandwidth of 2 TB/s.} in Figure \ref{fig:motivation:roofline}, \textbf{\texttt{W4A16}} has a higher theoretical throughput when $m < 78$, while \textbf{\texttt{W8A8}} performs better when $m > 78$. When the input batch size is small, GEMMs in LLMs are memory bound, and the memory bandwidth is dominated by weight traffic. Therefore, the smaller memory footprint of \textbf{\texttt{W4A16}} leads to better performance. However, when $m$ is large, the problem is compute bound. Thus, \textbf{\texttt{W8A8}} has faster speed thanks to the higher throughput from INT8 tensor cores. Intuitively, one can expect \textbf{\texttt{W4A8}} to combine the best of both worlds across all batch sizes. This is clearly demonstrated in Figure \ref{fig:motivation:roofline}, as long as we can perform all computation on INT8 tensor cores. 

Why \textbf{\texttt{KV4}}: attention workloads in LLM decoding can be formulated as a sequence of batched GEMV operations, with a computation intensity of 1 MAC / element regardless of input batch sizes. As in Equation \ref{eqn:background:attn}, the memory traffic is dominated by KV cache access, since $S \gg N = 1$ for each sequence. Quantizing the KV cache can be viewed as effectively increasing the memory bandwidth. Therefore, \textbf{\texttt{KV4}} offers 2$\times$ peak performance for attention over \textbf{\texttt{KV8}}. This improvement offers decent end-to-end speedup opportunities, since attention accounts for more than 50\% of total runtime at batch=64 in Figure \ref{fig:motivation:precision_choice}a. 

\subsection{Why Not \textbf{\texttt{W4A4KV4}}: Main Loop Overhead in GEMM}

\begin{figure}[t]
    \centering
    \includegraphics[width=\linewidth]{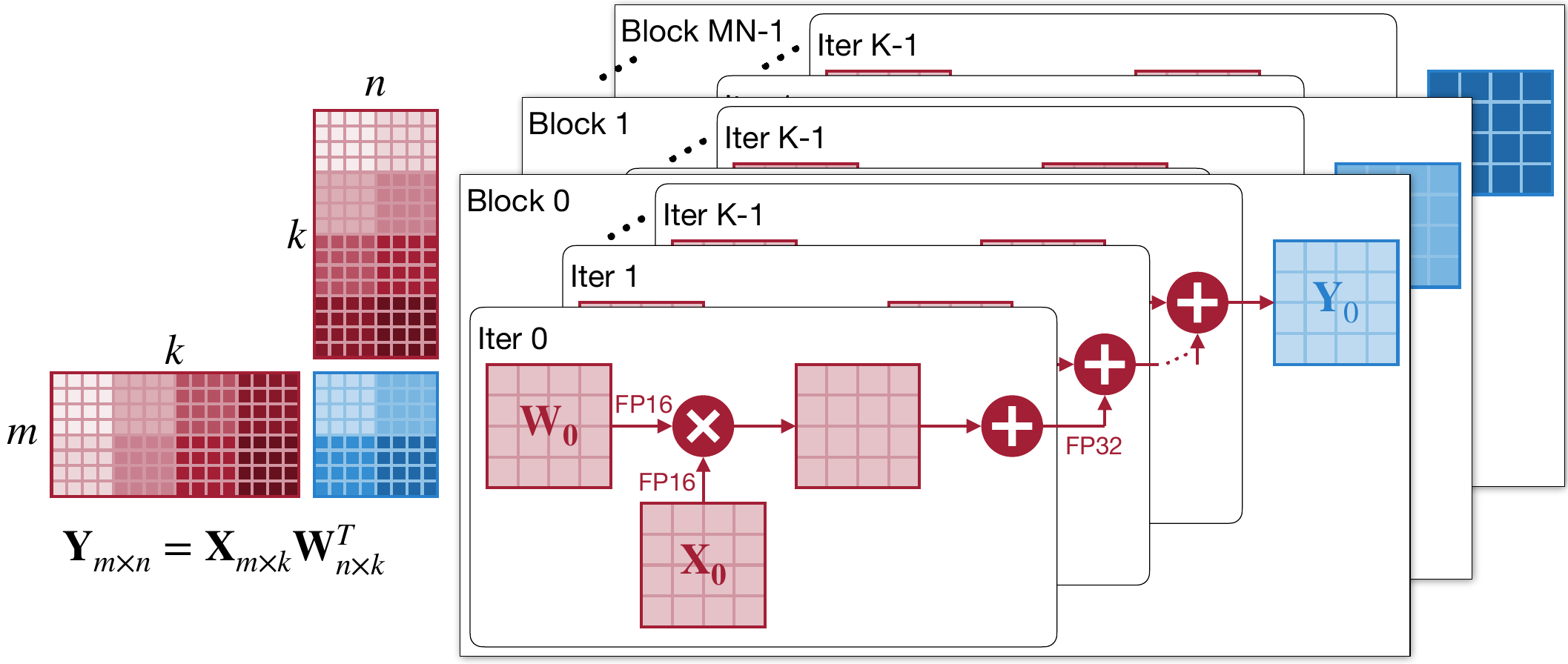}
    \caption{\textbf{Illustration of $m\times n\times k$ GPU GEMM}: $m, n$ are parallel dimensions and the reduction dimension $k$ has a sequential main loop. In LLM serving, $m$ is small and $n, k$ are large. Thus, the main loop is long.}
    \label{fig:background:gemm}
    \vspace{-20pt}
\end{figure}

\begin{figure*}[t]
    \centering
    \includegraphics[width=\linewidth]{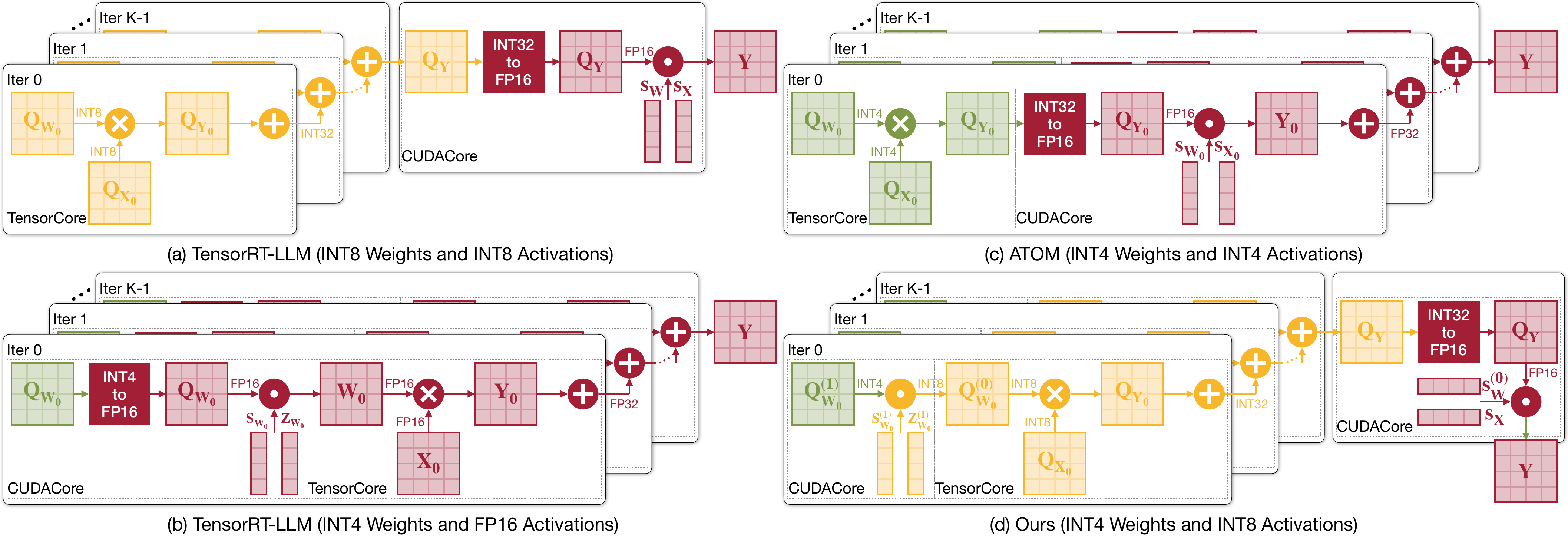}
    \caption{\textbf{Quantized GEMM on GPUs:} \textbf{\texttt{W8A8}} is fast because its main loop only contains \textit{tensor core} operations and all dequantization operations are present in the epilogue. Atom-\textbf{\texttt{W4A4}} and TensorRT-LLM-\textbf{\texttt{W4A16}} suffer from significant partial sum or weight dequantization overhead in the main loop. Thanks to the two-level progressive quantiation algorithm, \system-\textbf{\texttt{W4A8}} reduces main loop dequantization overhead by introducing register-level parallelism.}
    \label{fig:motivation:gemm}
    \vspace{-8pt}
\end{figure*}

\label{sect:motivation:dequantization-overhead}

A natural follow-up question would be: ``Why do we not choose the even more aggressive \textbf{\texttt{W4A4}}?" \textbf{\texttt{W4A4}} starts to achieve better theoretical GEMM performance when $m$, the number of input sequences, exceeds 78%
, as 4-bit tensor cores are twice as performant compared to their 8-bit counterparts. However, apart from the significant accuracy degradation, which will be discussed in Section \ref{sect:evaluation}, we demonstrate that such theoretical performance gains cannot be realized on existing GPU architectures (Ampere and Hopper). As in Figure \ref{fig:motivation:precision_choice}b, existing \textbf{\texttt{W4A4}} serving systems Atom~\cite{zhao2023atom} and QuaRot~\cite{ashkboos2024quarot} are even significantly slower than the \textbf{\texttt{W16A16}} solution from TensorRT-LLM. 

While this performance gap can be partially explained by the inefficient runtime in these two systems, the inherent difficulty in mapping  \textit{per-group quantized \textbf{\texttt{W4A4}} GEMM} on GPUs has been overlooked in previous literature. State-of-the-art systems implement tensor core GEMM with an output stationary dataflow shown in Figure \ref{fig:background:gemm}. For an $m\times n\times k$ problem, each thread block computes a $t_m\times t_n$ output tile by iterating sequentially through the reduction dimension $k$. This sequential loop is referred to as the main loop. The main loop comprises more than 100 iterations and dominates the runtime of the GEMM kernel. In both \textbf{\texttt{FP16}} and \textbf{\texttt{W8A8}} GEMM (Figure \ref{fig:motivation:gemm}a), the main loop is executed entirely on tensor cores. TensorRT-LLM-\textbf{\texttt{W4A16}} (Figure \ref{fig:motivation:gemm}b) and Atom-\textbf{\texttt{W4A4}} (Figure \ref{fig:motivation:gemm}c) both require dequantization operations in the main loop, which is running on the CUDA cores. \textbf{\texttt{W4A16}} requires \texttt{INT4} to \texttt{FP16} weight conversion, while Atom-\textbf{\texttt{W4A4}} requires \texttt{INT32} to \texttt{FP32} partial sum conversion and accumulation.

The dequantization process in Atom's main loop leads to two substantial efficiency bottlenecks. Firstly, on modern data center GPUs like the A100 and H100, the peak performance of \texttt{FP32} CUDA cores is merely \textbf{2\%} of their \texttt{INT4} tensor core counterparts. That said, de-quantizing one single partial sum in Atom is equivalent to \textbf{50} tensor core MACs. Therefore, the main loop is dominated by slow CUDA core operations rather than fast tensor core operations. Secondly, Atom creates two sets of registers (one for \texttt{FP32} and one for \texttt{INT32}) to hold partial sums. Larger GEMM problems (e.g., \contextstage) are typically register-bound on GPUs due to the nature of the output stationary dataflow, which results in high register consumption for storing \textit{partial sums}. Consuming a large number of registers within each warp limits the number of warps that can be executed simultaneously on the streaming multiprocessor. It is important to note that GPUs rely on low-cost context switching between a large number of in-flight warps to hide latency. Consequently, a smaller number of concurrently executed warps limits the opportunity for latency hiding, further exacerbating the main loop overhead.

We preview our \system's \textbf{\texttt{W4A8}} per-group quantized GEMM kernel design in Figure \ref{fig:motivation:gemm}d. We employ a two-level \textit{progressive group quantization} approach to ensure that all computations are performed on \texttt{INT8} tensor cores. We opt for weight dequantization over partial sum dequantization due to its lower register pressure. Furthermore, we apply 4-way \textit{register-level parallelism} to decode four \texttt{INT4} weights simultaneously, further reducing the main loop overhead.

\section{\algo Quantization}
\label{sect:algorithm}
To this end, we have discussed why \texttt{W4A8KV4} is a superior quantization precision choice. Yet, preserving model accuracy with such low-bit quantization remains a significant challenge. To unleash the full potential of  \texttt{W4A8KV4} without compromising the efficacy of large language models, we propose \algo algorithm featuring \textit{progressive group quantization}, \textit{\smoothattention}, and various general quantization optimizations. 

\subsection{Progressive Group Quantization}
\label{sect:alg:progressive-quant}

\begin{figure*}[t]
    \centering
    \includegraphics[width=\linewidth]{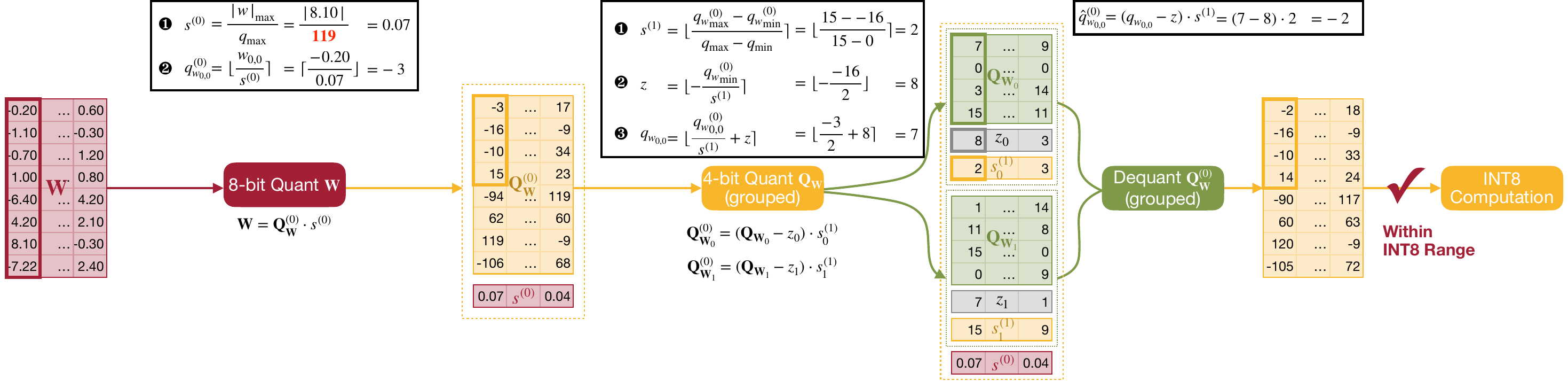}
    \vspace{-20pt}
    \caption{\textbf{Progressive Group Quantization} first employs per-channel INT8 quantization with protective range [-119, 119], followed by per-group INT4 quantization, so that the dequantized intermediate values remain within the INT8 range for computation. %
    }
    \label{fig:alg:progressive-quant}
    \vspace{-15pt}
\end{figure*}

To enhance the accuracy of low-bit quantization, \textit{group quantization} is commonly utilized~\cite{zhao2023atom,lin2023awq,gptq}. However, as outlined in Section~\ref{sect:motivation:dequantization-overhead}, the dequantization overhead in the system implementation can negate these accuracy improvements. To tackle this issue, we introduce progressive group quantization, as depicted in Figure~\ref{fig:alg:progressive-quant}.

Given the weight tensor $\mathbf{W} \in \mathbb{R}^{k \times n}$, we first apply \textit{per-channel} \textit{symmetric} \texttt{INT8} quantization:

\vspace{-10pt}
\begin{equation}
    \small \hat{\mathbf{W}} = {\mathbf{Q}_{\mathbf{W}}}^{(0)}_{\mathrm{s8}} \cdot \mathbf{s}^{(0)}_{\mathrm{fp16}},
    \label{eqn:alg:progressive_quant:int8}
\end{equation}
\vspace{-18pt}

where ${\mathbf{Q}_{\mathbf{W}}}_{\mathrm{s8}}^{(0)} \in \mathbb{N}^{n \times k}$ is the \textit{intermediate} 8-bit quantized weight tensor, and $\mathbf{s}^{(0)}_{\mathrm{fp16}} \in \mathbb{R}^{n \times 1}$ is the channel-wise quantization scales. We then further employ \textit{per-group} \textit{asymmetric} INT4 quantization on the intermediate weight tensor:

\vspace{-10pt}
\begin{equation}
 \small {{\mathbf{Q}}_{\mathbf{W}}}_\mathrm{s8}^{(0)} = \left({\mathbf{Q}_{\mathbf{W}}}_{\mathrm{u4}} - \mathbf{z}_{\mathrm{u4}} \right)\cdot \mathbf{s}^{(1)}_{\mathrm{u8}},
    \label{eqn:alg:progressive_quant:int4}
\end{equation}
\vspace{-18pt}

where ${\mathbf{Q}_{\mathbf{W}}}_{\mathrm{u4}} \in \mathbb{N}^{n \times k}$ is the unsigned 4-bit quantized weight tensor, $\mathbf{z}_{\mathrm{u4}} \in \mathbb{N}^{n \times k / g}$ is the unsigned 4-bit group-wise quantization zero points, and $\mathbf{s}^{(1)}_{\mathrm{u8}} \in \mathbb{N}^{n \times k/g}$ is the unsigned 8-bit group-wise quantization scales.

For \textbf{\texttt{W4A8}} GEMM computation, the 4-bit quantized weight tensor ${\mathbf{Q}_{\mathbf{W}}}_{\mathrm{u4}}$ will be first dequantized into intermediate 8-bit quantized weight tensor ${\mathbf{Q}_{\mathbf{W}}}_{\mathrm{s8}}^{(0)}$ following \eqn{eqn:alg:progressive_quant:int4}, and then perform \texttt{INT8} matrix multiplication as if it was \textbf{\texttt{W8A8}} per-channel quantization.

\paragraph{\textbf{Protective Quantization Range.}} Naively applying \eqn{eqn:alg:progressive_quant:int8} and~\ref{eqn:alg:progressive_quant:int4} does not guarantee that the intermediate dequantized weights perfectly lie in the 8-bit integer representation range (\ie, $[-128, 127]$). For example, after \texttt{INT8} quantization, a group of 8-bit weights lie in $[-113, 120]$. 4-bit asymmetric quantization will yield a scale factor of $\lceil (120--113)/(15-0) \rfloor =16$ and a zero point of $\lceil 0 - -113/16 \rfloor = 7$. Thus value 120 is quantized into $\lceil 120/16+7 \rfloor = 15$. It will be dequantized into $(15 - 7) * 16 = 128$ which is beyond the max 8-bit integer 127. One straightforward solution is to turn on the saturation option in the arithmetic instructions during dequantization. However, simply applying saturation will severely damage the computation throughput, reducing speed by as much as 67\%.

We reconsider the dequantization process. Take \eqn{eqn:background:quantization} into \eqn{eqn:alg:progressive_quant:int4}, we have,

\vspace{-18pt}
\begin{equation*}
    \small \hat{q}_{s8} = \lfloor \frac{{q}_{s8}}{{s}_{u8}} \rceil \cdot {{s}_{u8}} \le  {q}_{s8} + \frac{1}{2} {{s}_{u8}}
\end{equation*}
\vspace{-13pt}

Since ${s}_{u8} = \frac{{{q}_{s8}}_{\max} - {{q}_{s8}}_{\min}}{{{q}_{u4}}_{\max} - {{q}_{u4}}_{\min}} \le \frac{127-(-128)}{15-0} = 17$, we have,

\vspace{-18pt}
\begin{equation*}
    \small \hat{q}_{s8} \le 127  \rightarrow {q}_{s8} \le 127 - \frac{1}{2} {{s}_{u8}}  \rightarrow {q}_{s8} \le 119.5
\end{equation*}

\vspace{-10pt}
Therefore, we shrink the \texttt{INT8} symmetric quantization range from [-127, 127] to a \textit{protective range} [-119, 119] in order to avoid the dequantization overflow, as shown in the top of \fig{fig:alg:progressive-quant}.

\paragraph{\textbf{Compared to previous two-level quantization.}} Progressive group quantization introduces two levels of scales $\mathbf{s}^{(0)}_{\mathrm{fp16}}$ and $\mathbf{s}^{(1)}_{\mathrm{u8}}$. Prior studies such as VSQuant and DoubleQuant in QLoRA~\cite{dettmers2023qlora} also introduce two levels of scales to reduce the memory footprint of group-wise scaling factors. In contrast to our quantization flow, previous approaches directly apply group quantization with the target precision and then perform per-channel quantization on the group-wise floating-point scaling factors, as shown in the bottom of \fig{fig:alg:progressive-quant}:

\vspace{-12pt}
\begin{equation}
    \small \hat{\mathbf{W}} = {\mathbf{Q}_{\mathbf{W}}}_{\mathrm{s4}} \cdot \mathbf{s}_{\mathrm{fp16}}, \;\;\;\hat{\mathbf{s}}_{\mathrm{fp16}} = {\mathbf{s}}^{(1)}_{\mathrm{u8}} \cdot \mathbf{s}^{(0)}_{\mathrm{fp16}}
\end{equation}
\vspace{-15pt}

Therefore, using the group-wise scaling factors ${\mathbf{s}}^{(1)}_{\mathrm{u8}}$ to dequantize $\mathbf{Q}_{\mathbf{W}_{\mathrm{s4}}}$ cannot yield the 8-bit weight tensor. During the computation on GPUs, these approaches usually first dequantize the scales and, subsequently, the weights into floating-point values, which ultimately limits the peak throughput.

DGQ~\cite{zhang2023dual} also follows the quantization scheme of VSQuant and DoubleQuant, but enforces restrictions on scaling factors to make sure that all computation can be mapped onto \texttt{INT8} tensor cores. However, the DGQ serving system separates dequantization kernel with the GEMM kernel. Consequently, the end-to-end latency of \texttt{W4A8} GEMM in DGQ is even slower than the \texttt{W8A8} GEMM in cuBLAS, failing to demonstrate the memory bandwidth advantage of 4-bit weight quantization. In contrast, our \algo introduces a protective range, allowing us to fuse dequantization operations into the \texttt{W4A8} GEMM kernel with full register-level parallelism, minimizing CUDA core overhead. Thus, our \system's \texttt{W4A8} per-group GEMM achieves 1.5\x speedup over the \texttt{W8A8} cuBLAS GEMM.

\subsection{SmoothAttention}
\label{sect:alg:smooth-attention}
\begin{figure}[t]
    \centering
    \includegraphics[width=\linewidth]{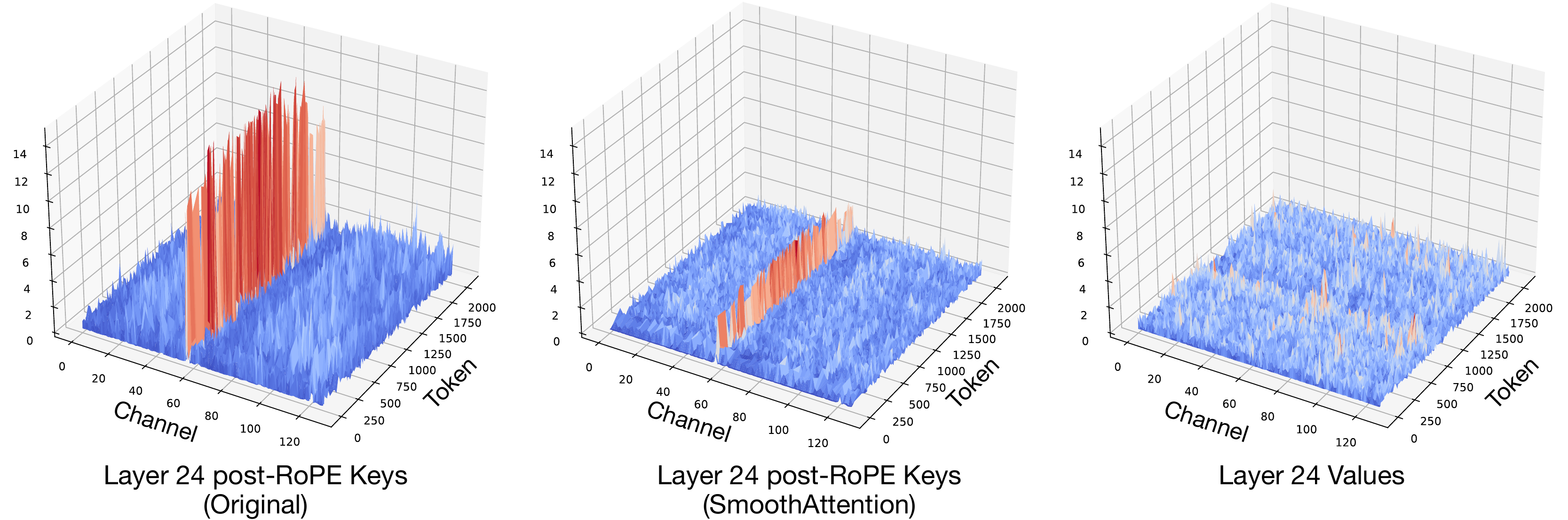}
    \caption{\textbf{SmoothAttention effectively smooths the outliers in Keys. Values doesn't suffer from outliers.}}
    \label{fig:alg:smooth-attention}
    \vspace{-10pt}
\end{figure}

As illustrated in \fig{fig:evaluation:alg-ablation}, directly reducing the KV cache to 4 bits significantly degrades the LLM accuracy. We visualize the magnitude distributions of the sampled Key and Value cache activations in \fig{fig:alg:smooth-attention}. We observe that: \textbf{the Value matrices show no significant outlier pattern, whereas Key matrices tend to have fixed outlier channels in each head}. These outliers are  $\sim$10\x larger than most of activation values. Though they can be easily handled \textbf{\texttt{KV8}} quantization in prior works~\cite{xiao2023smoothquant}, it places challenging obstacle to \textbf{\texttt{KV4}} quantization due to less quantization levels.

Inspired by SmoothQuant~\cite{xiao2023smoothquant}, we propose \smoothattention to scale down the outlier channels in Key cache by a per-channel factor $\mathbf{\lambda}$:

\vspace{-10pt}
\begin{equation}
    \small \mathbf{Z} = \left(\mathbf{Q}\mathbf{\Lambda}\right)\cdot \left(\mathbf{K}\mathbf{\Lambda}^{-1}\right)^T,\;\;\;\mathbf{\Lambda}=\mathrm{diag}\left(\mathbf{\lambda}\right)
\end{equation}
\vspace{-15pt}

SmoothQuant migrates the quantization difficulty from activations to weights, and thus requires a dedicate balance between activation and weight quantization by searching the migration strength. In contrast, since we do not quantize Queries, we only need to concentrate on the Keys and simply choose the \smoothattention scale factor as,

\vspace{-10pt}
\begin{equation}
    \small \mathbf{\lambda}_{i} = \max\left(|\mathbf{K}_i|\right)^{\alpha}.
\end{equation}
\vspace{-15pt}

In practice, $\alpha=0.5$ is good enough. As shown in \fig{fig:alg:smooth-attention}, after \smoothattention, the outliers in Key cache have been greatly smoothed.

In order to eliminate the extra kernel call overhead for \smoothattention scaling, fusing the scale into preceding linear layer's weights is preferred. However, modern LLMs employ the rotary positional embedding (RoPE) to both Keys and Queries, which needs extra handling. In practice, rotary positional embedding pairs channel $i$ with channel $i+\frac{D}{2}$ within each head. Consequently, to make \smoothattention scaling commutative in terms of RoPE, we add a hard constraint that $\lambda_{i} = \lambda_{i + \frac{D}{2}}$, and accordingly,

\vspace{-15pt}
\begin{equation}
   \small \mathbf{\lambda}_{i} = \lambda_{i + \frac{D}{2}} = \max\left(\max\left(|\mathbf{K}_i|\right), \max\left(|\mathbf{K}_{i+\frac{D}{2}}|\right)\right)^{\alpha}
\end{equation}
\vspace{-15pt}

Afterwards, we can easily fuse the \smoothattention scale $\mathbf{\Lambda}$ into previous layers' weights following $\mathbf{W}_{Q} = \mathbf{\Lambda}\mathbf{W}_{Q}$ and  $\mathbf{W}_{K} = \mathbf{\Lambda}^{-1}\mathbf{W}_{K}$.

\subsection{General LLM Quantization Optimizations}
\label{sect:alg:other}
One of the key challenges of low-bit LLM quantization is the activation outliers for every linear layers. We apply different optimizations for different types of linear layers as discussed below.

\begin{figure}[t]
    \centering
    \begin{minipage}[b]{\linewidth}
    \centering
    \includegraphics[width=\linewidth]{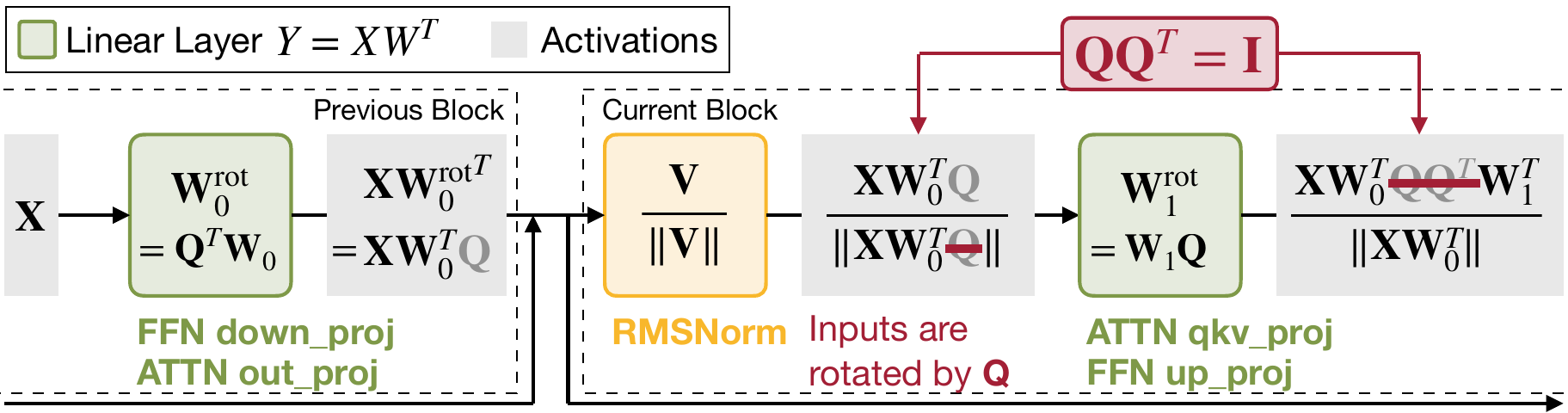}
    \caption{\textbf{Rotate the block input activations to suppress the outliers}: since rotation is a unitary transformation, the rotation matrix $\mathbf{Q}$ can be absorbed by the weights of the output module in the previous block.}
    \label{fig:alg:other:rotation}
    \end{minipage}
    \begin{minipage}[b]{\linewidth}
    \centering
    \includegraphics[width=\linewidth]{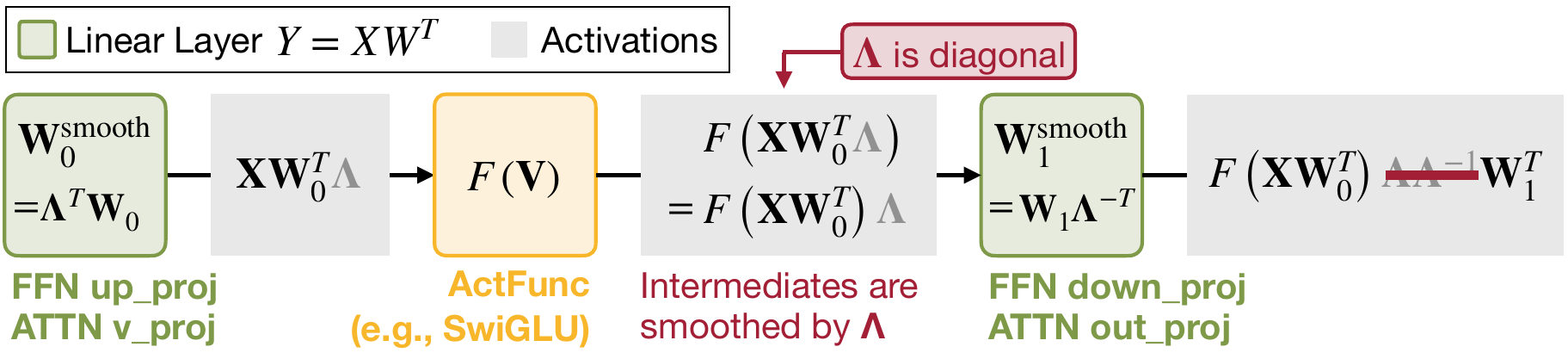}
    \caption{\textbf{Smooth the block intermediate activations, migrating the quantization difficulty to weights}: since smoothing is channel-independent, the smooth matrix $\mathbf{\Lambda}$ is diagonal and can be absorbed by the weights of the previous modules.}
    \label{fig:alg:other:smooth}
    \end{minipage}
\vspace{-45pt}
\end{figure}

\subsubsection{Block Input Module Rotation}
\label{sect:alg:other:rotation}
In transformer blocks, we define the components that take in the block inputs as input modules, such as the QKV Projection Layer and the FFN 1st Layer. 
As shown in \fig{fig:alg:other:rotation}, inspired by~\cite{ashkboos2024quarot,quip}, we rotate the block input activations by multiplying the rotation matrix. To keep mathematical equivalence of linear layers, we rotate the corresponding weights accordingly in the reversed direction.
After rotation, each channel's activations are linear combinations of all other channels, and thus outlier channels are effectively suppressed. Furthermore, since rotation is a unitary transformation, we can fuse the rotation matrix with the previous linear layers' weights. We simply choose the scaled Hadamard matrix as the rotation matrix.

\subsubsection{Block Output Module Smoothing}
\label{sect:alg:other:smooth}
Output modules refer to those layers that generate block outputs, such as the Output Projection Layer and FFN 2nd Layer. As shown in \fig{fig:alg:other:smooth}, inspired by~\cite{xiao2023smoothquant}, we smooth the block intermediate activations through dividing them by a per-channel smoothing factor. Original SmoothQuant does not smooth the block intermediate activations; moreover, if we directly smooth these modules with the same migration strength as input modules (\eg, \texttt{q\_proj}, \texttt{up\_proj}), the evaluated Wikitext-2 perplexity of the Llama-2-7B model will drop by as much as 0.05. In practice, we find that the migration strength $\alpha$ should be near 0. That is, the smoothing factor $\lambda$ is mostly determined by weights instead of activations, which is very different from the observations in SmoothQuant.

\subsubsection{Activation-Aware Channel Reordering}
\label{sect:alg:other:reorder}

\begin{figure}[t]
    \centering
    \includegraphics[width=\linewidth]{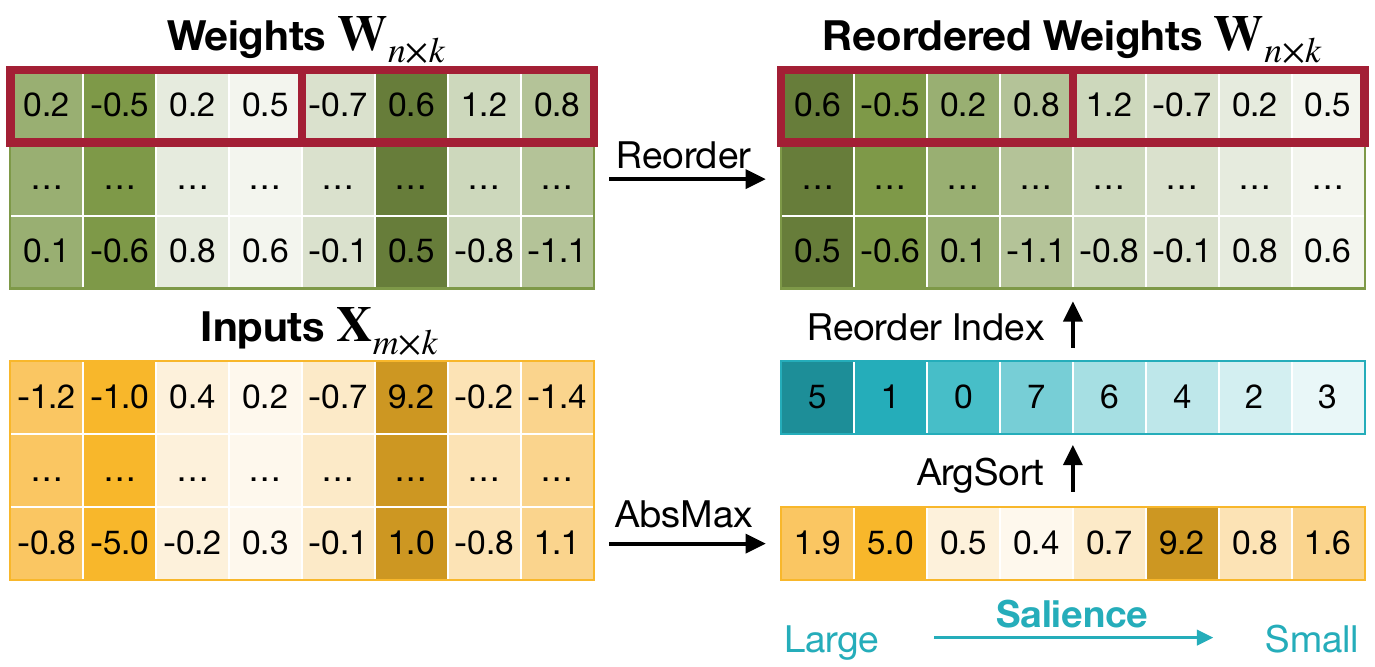}
    \caption{Reorder weight input channels based on their salience in group quantization. Channel salience can be determined by the magnitude of input activations.}
    \label{fig:alg:other:reorder}
\end{figure}

Both AWQ~\cite{lin2023awq} and Atom~\cite{zhao2023atom} have observed that maintaining the salient weights in FP16 can significantly improve model accuracy. These salient weights can be identified by the activation distribution. Instead of introducing mixed-precision quantization used by Atom, we propose activation-aware channel reordering as shown in \fig{fig:alg:other:reorder}. We use $\max\left(|\mathbf{X}|\right)$ to determine the channel salience, and then reorder channels so that channels with similar salience are in the same quantization group.

\subsubsection{Weight Clipping}
\label{sect:alg:other:clipping}
Weight clipping is another popular quantization optimization technique. It applies a clip ratio $\alpha$ to the dynamic range in \eqn{eqn:background:quantization} by letting $\mathbf{W}_{\max}=\alpha \max\left(\mathbf{W}\right)$ and $\mathbf{W}_{\min}=\alpha \min\left(\mathbf{W}\right)$.
Previous approaches~\cite{gptq,zhao2023atom,ashkboos2024quarot, lin2023awq} grid search the clip ratio $\alpha$ to minimize either quantization error of tensor itself (\ie, $\|\mathbf{W} - Q\left(\mathbf{W};\alpha\right)\|$) or output mean square error (\ie, $\|\mathbf{X}\mathbf{W}^T - \mathbf{X}Q\left(\mathbf{W}^T;\alpha\right)\|$. In \algo, we minimize the layer output error for all linear layers, expect for \texttt{q\_proj} and \texttt{k\_proj}, for which we optimize block output mean square error:

\vspace{-15pt}
\begin{equation}
    \small \arg\min_{\alpha} \|\mathrm{Block}\left(\mathbf{X}; \mathbf{W}\right) - \mathrm{Block}\left(\mathbf{X}; Q\left(\mathbf{W}; \alpha\right)\right)\|.
\end{equation}

\section{\system Serving System}
\label{sect:system}

To this end, we have presented the \algo quantization algorithm, which aims to minimize accuracy loss incurred by \textbf{\texttt{W4A8KV4}} quantization. However, realizing the theoretical throughput benefits in Figure \ref{fig:motivation:roofline} remains challenging. Thus, in this section, we will delve into the \system system design, which follows two important directions: \textbf{I.} Reducing main loop overhead in GEMM kernels; \textbf{II.} Accerating KV4 attention. %

\subsection{\system System Runtime}
\label{sect:system:runtime}

\begin{figure}[t]
    \centering
    \includegraphics[width=0.85\linewidth]{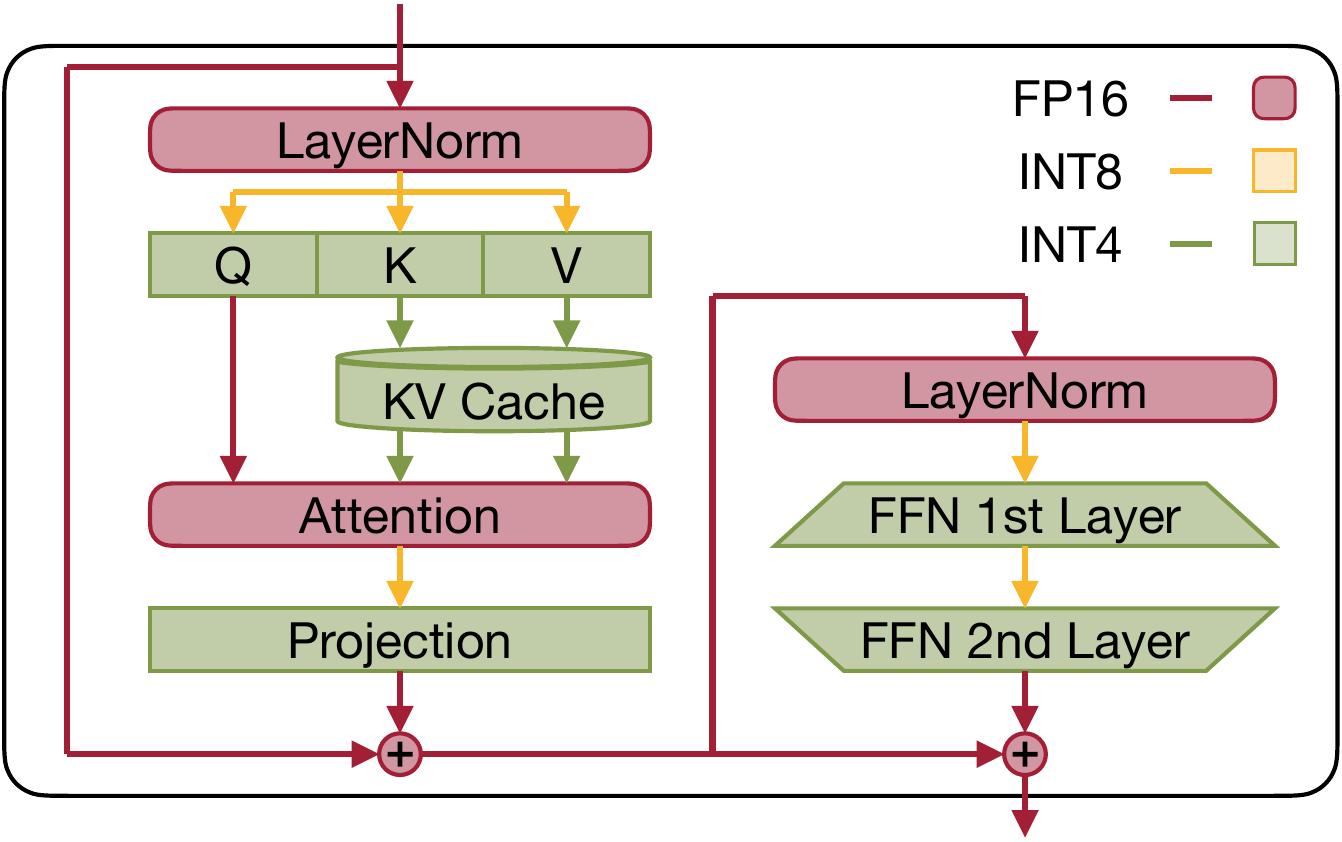}
    \caption{\textbf{\system's precision mapping} for an \texttt{FP16} in, \texttt{FP16} out LLM block. All GEMM operators take in \textbf{\texttt{W4A8}} inputs and produce \texttt{FP16} outputs. Activation quantization happens in normalization and activation layers.}
    \label{fig:system:overview}
    \vspace{-10pt}
\end{figure}

We start by introducing the \system runtime in Figure \ref{fig:system:overview}. All GEMM layers in \system operate on \textbf{\texttt{W4A8}} inputs, perform computation on \texttt{INT8} tensor cores, and generate \texttt{FP16} outputs. 
All attention layers perform computation in \texttt{FP16} on CUDA cores. Consequently, each LLM block in \system has \texttt{FP16} inputs and \texttt{FP16} outputs. 

\textbf{Activation Quantization.} To ensure that each GEMM takes in \texttt{INT8} activation, we fuse activation quantization into the preceding layernorm for the QKV projection and the first FFN layer, or into the preceding activation kernel for the second FFN layer. Furthermore, a separate quantization node is inserted before output projection in the attention block.

\textbf{KV Cache Management.} To avoid memory fragmentation, we follow vLLM~\cite{kwon2023efficient} and TensorRT-LLM~\cite{trtllm} to adopt paged KV caches. In contrast to these two frameworks, which perform \textit{per-tensor}, \textit{static} quantization (\ie, scaling factors computed offline) on KV caches, \system requires \textit{per-head}, \textit{dynamic} KV quantization to maintain competitive accuracy due to the lower bit precision (4 \vs 8). We therefore store \texttt{FP16} scaling factors and zero points for each head immediately following the quantized KV features in each KV cache page, allowing these values to be updated on-the-fly. \system also supports in-flight batching, similar to vLLM and TensorRT-LLM. %

\subsection{\textbf{\texttt{W4A8}} GEMM in \system}

As discussed in Section \ref{sect:motivation}, the main loop overhead poses a significant obstacle in allowing quantized GEMMs to attain the theoretical performance gains projected by the roofline model (Figure \ref{fig:motivation:roofline}). Therefore, the focus of \system \textbf{\texttt{W4A8}} GEMM is to \textbf{reduce main loop overhead}. Specifically, we address the costs of pointer arithmetic operations through \textbf{compute-aware weight reorder}, and reduce dequantization overhead through a \textbf{subtraction after multiplication} computation order and \textbf{register-level parallelism}.

\subsubsection{Compute-Aware Weight Reorder}

\begin{figure}[t]
    \centering
    \includegraphics[width=\linewidth]{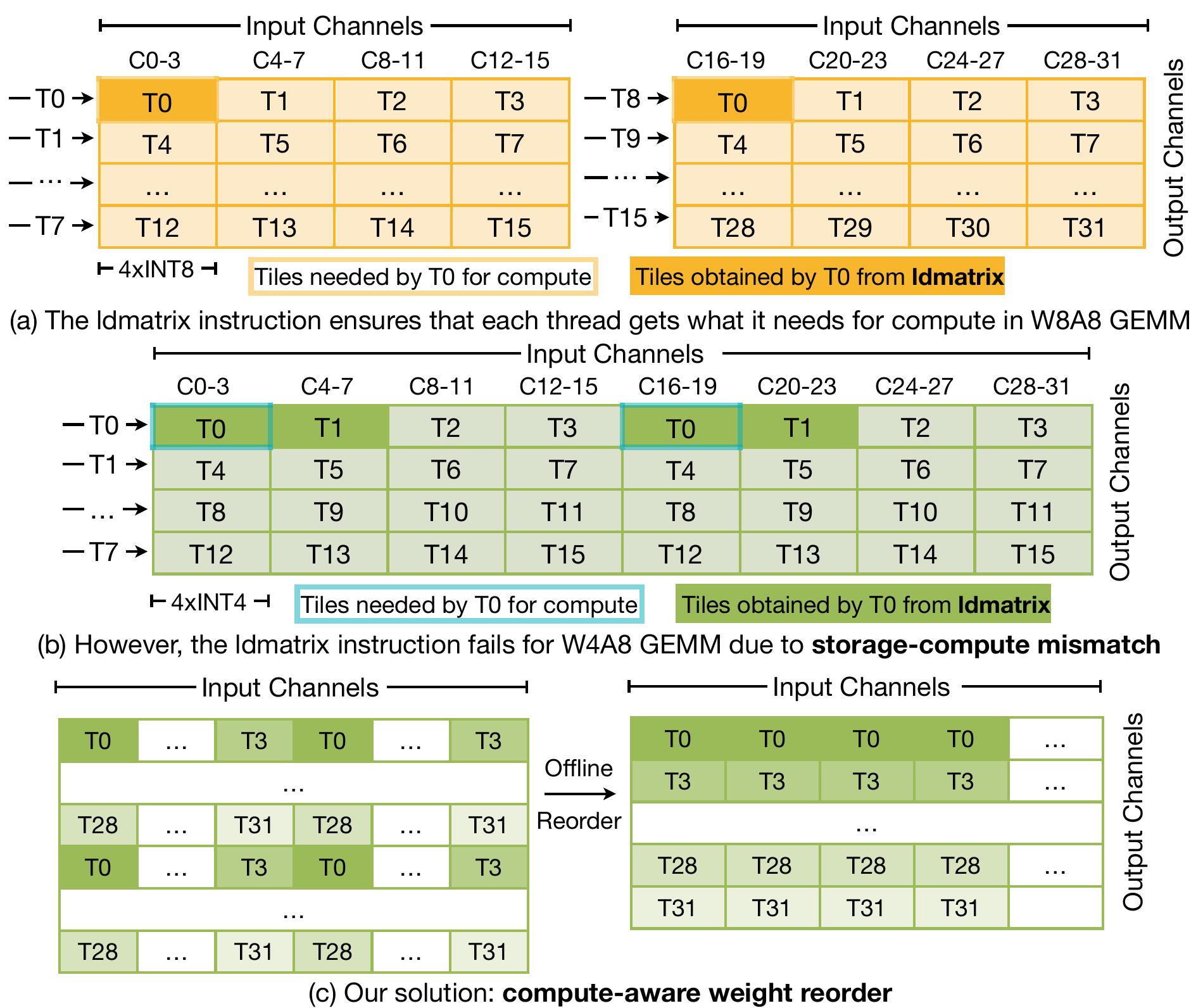}
    \vspace{-15pt}
    \caption{\system applies \textbf{compute-aware weight reoder} to minimize the pointer arithmetics in \textbf{\texttt{W4A8}} GEMM main loop.} %
    \label{fig:sys:weight_reorder}
    \vspace{-20pt}
\end{figure}

Prior to dequantization and tensor core computation, the operands must be loaded from global memory into the L1 shared memory during each main loop iteration. This loading process is non-trivial since the tensor core GEMM intrisics require a strided layout for each thread in computation, as demonstrated in Figure \ref{fig:sys:weight_reorder}a. For instance, instead of loading consecutive eight \texttt{INT8} weights, thread 0 first loads input channels 0-3, then skips ahead to input channels 16-19. That said, a naive weight loading implementation would require one address calculation per four channels, leading to two efficiency issues. First, pointer arithmetic operations are performed on CUDA cores, which have \textbf{32\x} lower throughput than the \texttt{INT8} tensor core on the A100. Consequently, the address calculation overhead becomes non-negligible. Second, strided memory access prevents achieving the highest HBM bandwidth through packed 128-bit loading, further slowing down the memory pipeline. This issue is addressed by the \texttt{ldmatrix} instruction when the storage and compute data types are \textbf{the same}. As illustrated in Figure \ref{fig:sys:weight_reorder}a, thread $i$ loads a consecutive 128 bits in output channel $i \% 8$, and the \texttt{ldmatrix} instruction automatically distributes the data in a strided manner, ensuring that each thread eventually obtains the required data for \texttt{INT8} tensor core computation. 

Unfortunately, the \texttt{ldmatrix} instruction will \textbf{not} work when the data types used for storage and computation differ (like in \textbf{\texttt{W4A8}}). Specifically, in Figure \ref{fig:sys:weight_reorder}b, \texttt{ldmatrix} ensures that each thread obtains the same number of \textbf{bytes}, not the same number of \textbf{elements}, after data permutation in the register file. Consequently, thread 0 obtains the tiles needed by both itself and thread 1, while thread 1 obtains the tiles needed by thread 2 and thread 3 in the subsequent \texttt{INT8} tensor core computation. This creates a mismatch between the data obtained by each thread and used in computation. That said, \texttt{ldmatrix} cannot be used for \textbf{\texttt{W4A8}} GEMM and the aforementioned pointer arithmetic overhead persists. Worse still, memory bandwidth utilization deteriorates further as we consecutively load only 16 bits for 4-bit weights.

We address this challenge through \textbf{compute-aware weight reordering} (Figure \ref{fig:sys:weight_reorder}c). The key insight is to store the weights in the order they are used during computation. We divide the entire GEMM problem into multiple 32$\times$32 tiles. Within each tile, thread 0 utilizes input channels 0-3 and 16-19 for output channels 0, 8, 16, and 24 (output channels 16-31 are omitted in Figure \ref{fig:sys:weight_reorder}c). Consequently, we concatenate these 32 channels into a single 128-bit word. The 32 channels used by thread 1 are stored immediately following thread 0's 32 channels. Since weights are static, such reordering does not introduce any runtime overhead. Additionally, it not only reduces the pointer arithmetic overhead to the same level as \texttt{ldmatrix} but also guarantees high-bandwidth 128-bit/thread memory transactions. We apply this reordering to zero points and scales as well to mitigate dequantization overhead. 

\subsubsection{Fast Dequantization in Per-Channel \textbf{\texttt{W4A8}} GEMM}
\label{sect:system:gemm:per_channel}
\begin{figure}[t]
    \centering
    \includegraphics[width=\linewidth]{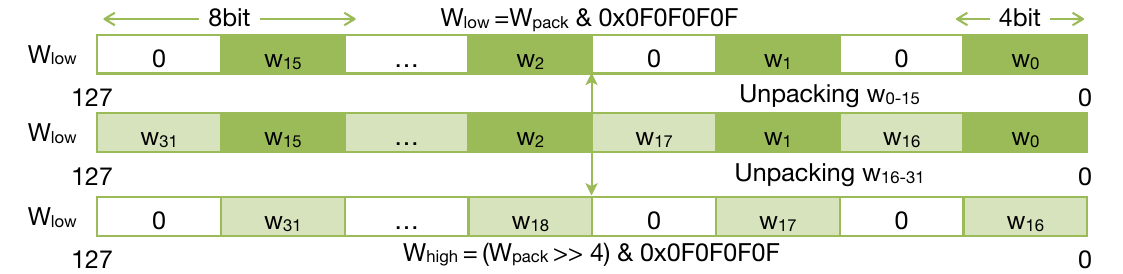}
    \caption{\system exploits \textbf{register-level parallelism} to significantly reduce the number of required logical operations in \texttt{UINT4} to \texttt{UINT8} weight unpacking.}
    \label{fig:sys:dequant_u4tou8}
    \vspace{-10pt}
\end{figure}

As illustrated in Figure \ref{fig:motivation:gemm}d, dequantizing weights within the main loop becomes necessary when the bit precisions for weights and activations differ. In the case of per-channel \textbf{\texttt{W4A8}} quantization, second-level scaling factors are omitted, and first-level \texttt{FP16} scaling is efficiently fused into the GEMM epilogue. We therefore focus our discussion on the efficient conversion from \texttt{ZINT4} (i.e., unsigned 4-bit integers with zero points) to \texttt{SINT8} within the main loop. We further decompose this conversion into two steps: \texttt{UINT4} to \texttt{UINT8} (weight unpacking) and \texttt{UINT8} to \texttt{SINT8} (zero point subtraction). As depicted in Figure \ref{fig:sys:dequant_u4tou8}, we reorder every 32 \texttt{UINT4} weights $w_0, w_1, ..., w_{31}$ into $w_0, w_{16}, w_1, w_{17}, ...$ This allows us to exploit \textbf{register-level parallelism} and efficiently unpack them into \texttt{UINT8} numbers with only three logical operations.

For the conversion from \texttt{UINT8} to \texttt{SINT8}, the most intuitive approach is to introduce integer subtraction instructions within the main loop, which we refer to as subtraction before multiplication. Although straightforward, this approach inevitably introduces additional cost to the main loop, which is undesirable. Instead, we adopt a \textbf{subtraction after multiplication} approach to minimize the main loop overhead.

Specifically, a GEMM layer with per-channel quantized operands can be expressed as:

\vspace{-17pt}
\begin{equation}
\small \mathbf{O} = \hat{\mathbf{X}}\hat{\mathbf{W}} = (\mathbf{Q}_\mathbf{X}\odot\mathbf{S}_\mathbf{X})((\mathbf{Q}_\mathbf{W} - \mathbf{Z}_\mathbf{W})\odot\mathbf{S}_\mathbf{W}),  
\label{eqn:sys:per_channel_qmm}
\end{equation}

\vspace{-10pt}
where $\mathbf{Q}_\mathbf{W}$ ($\mathbf{Q}_\mathbf{X}$) is the quantized weight (activation), $\mathbf{Z}_\mathbf{W}$ expands the zero point vector $\mathbf{z}_\mathbf{W}$ of size $n$ (output channels) to $k\times n$ ($k$ is input channels) and $\mathbf{S}_\mathbf{W}$, $\mathbf{S}_\mathbf{X}$ are similarly obtained from scaling vectors $\mathbf{s}_\mathbf{W},\mathbf{s}_\mathbf{X}$. We denote $\mathbf{Z}_\mathbf{W}\odot\mathbf{S}_\mathbf{W}$ as $\mathbf{ZS}_\mathbf{W}$, then we rewrite Equation \ref{eqn:sys:per_channel_qmm} as:

\vspace{-20pt}
\begin{equation}
\begin{aligned}
\small \mathbf{O} &= (\mathbf{Q}_\mathbf{X}\odot\mathbf{S}_\mathbf{X})(\mathbf{Q}_\mathbf{W}\odot\mathbf{S}_\mathbf{W}-\mathbf{ZS}_\mathbf{W}) \\
\small &= (\mathbf{Q}_\mathbf{X}\mathbf{Q}_\mathbf{W})\odot(\mathbf{s}_\mathbf{W}\times\mathbf{s}_\mathbf{X}) - (\mathbf{Q}_\mathbf{X}\odot\mathbf{S}_\mathbf{X})\mathbf{ZS}_\mathbf{W}.  
\label{eqn:sys:per_channel_qmm_step2}
\end{aligned}
\end{equation}

The first term, $(\mathbf{Q}_\mathbf{X}\mathbf{Q}_\mathbf{W})\odot(\mathbf{s}_\mathbf{W}\times\mathbf{s}_\mathbf{X})$, is analogous to the \textbf{\texttt{W8A8}} GEMM in TensorRT-LLM, where the $\mathbf{s}_\mathbf{W}\times\mathbf{s}_\mathbf{X}$ outer product scaling is performed in the epilogue. For the second term, we first replace $\mathbf{Q}_\mathbf{X}\mathbf{S}_\mathbf{X}$  ($\hat{\mathbf{X}}$) with the unquantized $\mathbf{X}$. We then notice that:

\vspace{-14pt}
\begin{equation}
\mathbf{X}(\mathbf{ZS}_\mathbf{W}) = \mathbf{t}_\mathbf{X}\times(\mathbf{z}_\mathbf{W}\odot\mathbf{s}_\mathbf{W}),
\label{eqn:sys:per_channel_qmm_step3}
\end{equation}
\vspace{-18pt}

where $\mathbf{t}_\mathbf{X} = \mathbf{X}\mathbf{1}_k$, i.e., summing all input channels for each token. We observe that Equation \ref{eqn:sys:per_channel_qmm_step3} has a form similar to the outer product of scaling factors. Therefore, it can also be fused into the epilogue of the \textbf{\texttt{W4A8}} GEMM, analogous to the first term in Equation \ref{eqn:sys:per_channel_qmm_step2}. To this end, we move the zero-point subtraction from the main loop to the epilogue, thereby largely eliminating its overhead in the GEMM kernel. This formulation of \textbf{subtraction after multiplication} necessitates precomputing $\mathbf{t}_\mathbf{X}$. Fortunately, each \textbf{\texttt{W4A8}} kernel is always preceded by a memory-bound kernel, allowing us to fuse the precomputation kernel into it with negligible latency overhead.

\subsubsection{Fast Dequantization in Per-Group \textbf{\texttt{W4A8}} GEMM} 
\begin{figure}[t]
    \centering
    \includegraphics[width=\linewidth]{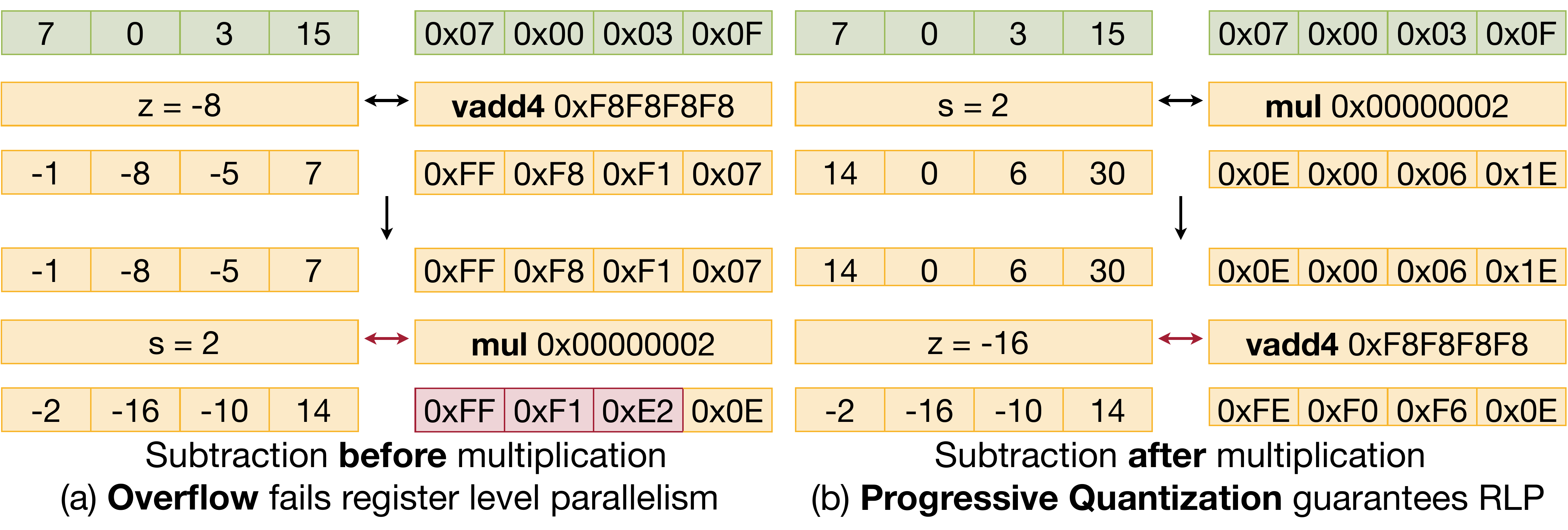}
    \caption{Our progressive quantization algorithm ensures that all intermediate results in the \textbf{subtraction after multiplication} computation order will not overflow, thereby enabling \textbf{register-level parallelism} and reducing main loop overhead.}
    \label{fig:sys:dequant_group}
    \vspace{-30pt}
\end{figure}

The primary distinction between the per-group \textbf{\texttt{W4A8}} GEMM and its per-channel counterpart lies in the second-level dequantization process in Figure \ref{fig:motivation:gemm}d. Firstly, since zero points are now defined on a per-group basis, it is no longer possible to merge zero point subtraction into the epilogue, as was done in the previous section. Secondly, due to the presence of level 2 scales, an additional \texttt{INT8} multiplication is required for each weight. Akin to the previous section, we must determine whether to apply multiplication (scales) or subtraction (zeros) first during level 2 dequantization.

In this context, we contend that performing \textbf{subtraction after multiplication} remains the advantageous approach because it enables \textbf{register-level parallelism} (RLP). As shown in Figure \ref{fig:sys:dequant_group}, NVIDIA GPUs provide the \texttt{vadd4} instruction that performs four \texttt{INT8} additions with a single \texttt{INT32} ALU operation. However, there is no instruction that realizes similar effect for 4-way \texttt{INT8} multiplication. Consequently, in order to achieve RLP, one has to simulate this by padding 24 zeros to the most significant bits (MSBs) of the 8-bit scaling factor. However, this simulation is valid only when the result of each \texttt{INT8} multiplication remains within the \texttt{INT8} range. This condition is not met for the subtraction-before-multiplication computation order. As illustrated in Figure \ref{fig:sys:dequant_group}a, the result of the scale multiplication overflows, leading to an incorrect output. In the subtraction-before-multiplication approach, we can only perform multiplication one by one, which is extremely inefficient. On the other hand, with the subtraction-after-multiplication computation order, our progressive group quantization algorithm ensures that the result of the initial multiplication step never exceeds the \texttt{INT8} range. This allows for fully leveraging the performance benefits of RLP in both multiplication and subtraction.

\vspace{-8pt}
\subsubsection{General Optimizations}

In our \textbf{\texttt{W4A8}} kernel, we also employ general techniques for GEMM optimization. On the memory side, we apply multi-stage software pipelining and asynchronous memory copy to better overlap memory access with computation. Additionally, we swizzle the layout of the L1 shared memory to eliminate bank conflicts. To improve L2 cache utilization, we permute the computation partition across different thread blocks, allowing adjacent blocks to reuse the same weight. On the compute side, when the number of input tokens ($m$) is small, we found it beneficial to partition the reduction dimension $k$ into multiple slices and reduce the partial sums across different warps in the L1 shared memory. %

\vspace{-8pt}
\subsection{\textbf{\texttt{KV4}} Attention in \system}
\label{sect:system:attention}

\begin{table}
\centering
\small
\caption{A naive \texttt{KV4} attention implementation is 1.7\x faster on L40S than TRT-LLM-\texttt{KV8}, but is 1.1-1.2\x slower on A100 due to earlier CUDA core roofline turning point.}
\scalebox{1.0}{
\begin{tabular}{cccc}
\toprule
Seq\_len & 8-bit KV & 4-bit KV (Naive) & 4-bit KV (Ours) \\
\midrule
128 & 0.09 ms & 0.10 ms \textcolor{red}{(0.87$\times$)} & 0.07 ms \textcolor{green!50!black}{(1.29$\times$)}\\
256 & 0.14 ms & 0.16 ms \textcolor{red}{(0.86$\times$)} & 0.11 ms \textcolor{green!50!black}{(1.32$\times$)}\\
512 & 0.23 ms & 0.27 ms \textcolor{red}{(0.87$\times$)} & 0.16 ms \textcolor{green!50!black}{(1.44$\times$)}\\
1024 & 0.42 ms & 0.48 ms \textcolor{red}{(0.88$\times$)} & 0.28 ms \textcolor{green!50!black}{(1.49$\times$)}\\
1536 & 0.62 ms & 0.69 ms \textcolor{red}{(0.90$\times$)} & 0.41 ms \textcolor{green!50!black}{(1.51$\times$)}\\
 \bottomrule
\end{tabular}
}
\label{tab:sys:naive_attn_slower}
\vspace{-20pt}
\end{table}

Attention accounts for 30-50\% of the total LLM runtime, as depicted in Figure \ref{fig:motivation:precision_choice}a. Although the roofline model in Figure \ref{fig:motivation:gemm} suggests that quantizing the KV cache to \texttt{INT4} should automatically yield a 2\x speedup over the 8-bit KV baseline, this is not the case in real-world implementation. 

We start with the \textbf{\texttt{KV8}}-attention \decodingstage kernel from TensorRT-LLM as our baseline and replace all static, per-tensor quantized 8-bit KV cache accesses and conversions with their dynamic, per-head quantized 4-bit counterparts. This direct replacement immediately leads to 1.7$\times$ speedup on L40S, but results in \textbf{1.2\x slowdown} on A100 (Table \ref{tab:sys:naive_attn_slower}), compared to the \textbf{\texttt{KV8}} baseline. 

Once again, our analysis reveals that the devil is in the slow CUDA cores, which are responsible for executing the attention kernels during the \decodingstage. While each individual batched GEMV has a computation intensity of 1 MAC / element, the computation intensity escalates significantly for a fused attention kernel that combines all the arithmetics and KV cache updates. As an illustration, naively dequantizing a single \texttt{INT4} number from the KV cache necessitates 5 ALU Ops. This includes \texttt{mask} and \texttt{shift} operations to isolate the operand, type conversion from integer to floating-point representation, and floating point \texttt{mul} and \texttt{sub} to obtain the final results. It is crucial to note that the roofline turning point for A100 FP32 CUDA cores is merely \textbf{9.8 Ops/Byte}. That said, the dequantization of KV  operands alone already saturates this bound, leading to the surprising observation that the fused \textbf{\texttt{KV4}} attention kernel can become \textbf{compute-bound} on datacenter GPUs like A100. In fact, similar observations hold in other systems like QuaRot~\cite{ashkboos2024quarot} and Atom~\cite{zhao2023atom}. Specifically, QuaRot introduces compute-intensive Hadamard transformation~\cite{quip} in the attention operator, making it hard to achieve real speedup over TRT-LLM-\texttt{KV8} with 4-bit quantized KV caches.

To mitigate the compute-bound bottleneck, it is important to shift the \decodingstage \textbf{\texttt{KV4}} attention kernels away from the compute-bound region. We accomplish this objective through a bidirectional approach: Firstly, delaying the onset of the roofline turning point, and secondly, concurrently reducing the computation intensity within the fused kernel. For the first part, we replace all FP32 operations in the original TensorRT-LLM kernel with their FP16 counterpart, effectively doubling the computation roof.  For the second part, we observe that the arithmetic intensity of dequantization can be significantly reduced to 2 operations per element by applying bit tricks proposed in~\cite{kim2022says}. Furthermore, we note that simplifying the control logic and prefetching the scaling factors and zero values, thereby simplifying address calculations, contribute to performance improvements. After incorporating these enhancements, we observe a \textbf{1.5\x} speedup over TensorRT-LLM's \textbf{\texttt{KV8}} kernel on A100.

\begin{table*}[htbp]
\centering
\small
\caption{WikiText2 perplexity with 2048 sequence length. The lower is the better.}
\scalebox{0.9}{
\begin{tabular}{llcccccccccc}
\toprule
  \multicolumn{2}{c}{WikiText2 Perplexity ↓} 
& Llama-3
& \multicolumn{3}{c}{Llama-2} 
& \multicolumn{3}{c}{Llama} 
& Mistral & Mixtral & Yi \\
\cmidrule{1-9}
  Precision         & Algorithm  
& 8B      
&   7B    &   13B   &   70B   
&   7B    &   13B   &   30B   
&   7B    &   8x7B  &   34B \\
\midrule
\texttt{FP16}       & - 
&  6.14   
&  5.47   &  4.88   &  3.32   
&  5.68   &  5.09   &  4.10   
&  5.25   &  3.84   &  4.60 \\
\midrule
\texttt{W8A8}       & SmoothQuant 
&  6.28   
&  5.54   &  4.95   &  3.36 
&  5.73   &  5.13   &  4.23   
&  5.29   &  3.89   &  4.69 \\
\midrule
\multirow{2}{*}{\makecell[l]{\texttt{W4A16} \\ g128}} & GPTQ-R 
&  6.56   
&  5.63   &  4.99   &  3.43
&  5.83   &  5.20   &  4.22   
&  5.39   &  4.08   &  4.68 \\
                    & AWQ 
&  6.54   
&  5.60   &  4.97   &  3.41
&  5.78   &  5.19   &  4.21   
&  5.37   &  4.02   &  4.67 \\
\midrule
\multirow{2}{*}{\texttt{W4A4}} & \multirow{2}{*}{QuaRot}
&  \textcolor{gray}{8.20}
&  \textcolor{gray}{6.10}   &  \textcolor{gray}{5.40}   &  \textcolor{gray}{3.79}
&  \textcolor{gray}{6.26}   &  \textcolor{gray}{5.55}   &  \textcolor{gray}{4.60}   
&  \textcolor{gray}{5.71}   &  \textcolor{gray}{NaN}    &  \textcolor{gray}{NaN} \\
              &  
&  8.33   
&  6.19   &  5.45   &  3.83
&  6.34   &  5.58   &  4.64   
&  5.77   &  NaN    &  NaN \\
\midrule
\multirow{4}{*}{\makecell[l]{\texttt{W4A4} \\ g128}} & \multirow{2}{*}{QuaRot}$\dagger$
&  \textcolor{gray}{7.32}
&  \textcolor{gray}{5.93}   &  \textcolor{gray}{5.26}   &  \textcolor{gray}{3.61}
&  \textcolor{gray}{6.06}   &  \textcolor{gray}{5.40}   &  \textcolor{gray}{4.44}   
&  \textcolor{gray}{5.54}   &  \textcolor{gray}{NaN}    &  \textcolor{gray}{NaN} \\
              &  
&  7.51   
&  6.00   &  5.31   &  3.64
&  6.13   &  5.43   &  4.48   
&  5.58   &  NaN    &  NaN \\
\cmidrule{2-12}
& \multirow{2}{*}{Atom}$\dagger$
&  \textcolor{gray}{7.57}
&  \textcolor{gray}{6.03}   &  \textcolor{gray}{5.27}   &  \textcolor{gray}{3.69}
&  \textcolor{gray}{6.16}   &  \textcolor{gray}{5.46}   &  \textcolor{gray}{4.55}   
&  \textcolor{gray}{5.66}   &  \textcolor{gray}{4.42}    &  \textcolor{gray}{4.92} \\
              &  
&  7.76   
&  6.12   &  5.31   &  3.73 
&  6.25   &  5.52   &  4.61   
&  5.76   &  4.48   &  4.97 \\
\arrayrulecolor{black}
\midrule
\multirow{3}{*}{\texttt{W4A8KV4}} & RTN
&  9.50   
&  6.51   &  5.40   &  3.90 
&  6.51   &  5.71   &  4.91   
&  6.18   &  5.02   &  6.52 \\
                    & AWQ
&  7.90   
&  6.28   &  5.25   &  3.68
&  6.33   &  5.59   &  4.61   
&  5.92   &  4.58   &  5.26 \\
                    & \cellcolor{mylightblue}\algo
& \cellcolor{mylightblue}\textbf{6.81}
&  \cellcolor{mylightblue}\textbf{5.75}   &  \cellcolor{mylightblue}\textbf{5.11}   & \cellcolor{mylightblue}\textbf{3.50}
&  \cellcolor{mylightblue}\textbf{5.92}   &  \cellcolor{mylightblue}\textbf{5.27}   & \cellcolor{mylightblue}\textbf{4.31}   
&  \cellcolor{mylightblue}\textbf{5.44}   &  \cellcolor{mylightblue}\textbf{4.17}   & \cellcolor{mylightblue}\textbf{4.73} \\
\midrule
\multirow{3}{*}{\makecell[l]{\texttt{W4A8KV4} \\ g128}} & RTN
&  7.25   
&  5.99   &  5.19   &  3.70
&  6.23   &  5.46   &  4.56   
&  5.59   &  4.39   &  5.49 \\
                    & AWQ
&  6.94   
&  5.83   &  5.12   &  3.51
&  5.93   &  5.36   &  4.39   
&  5.50   &  4.23   &  4.78 \\
                    & \cellcolor{mylightblue}\algo
& \cellcolor{mylightblue}\textbf{6.70}
& \cellcolor{mylightblue}\textbf{5.67} & \cellcolor{mylightblue}\textbf{5.06} & \cellcolor{mylightblue}\textbf{3.46}
& \cellcolor{mylightblue}\textbf{5.88} & \cellcolor{mylightblue}\textbf{5.23} & \cellcolor{mylightblue}\textbf{4.27}   
& \cellcolor{mylightblue}\textbf{5.41}          & \cellcolor{mylightblue}\textbf{4.13} & \cellcolor{mylightblue}\textbf{4.73}\\
\bottomrule
\multicolumn{11}{l}{* Grayed results use Wikitext2 as calibaration dataset.} \\
\multicolumn{11}{l}{$\dagger$ QuaRot and Atom apply group quantization to activations as well.} \\

\end{tabular}
}
\label{tab:evaluation:wikitext}
\end{table*}

\begin{table*}[t]
\centering
\small
\caption{Zero-shot accuracy on five common sense tasks with 2048 sequence length.}
\scalebox{0.98}{
\begin{tabular}{lllcccccc}
\toprule
\multirow{2}{*}{Llama-2} 
& \multirow{2}{*}{Precision} 
& \multirow{2}{*}{Method}
& \multicolumn{6}{c}{Zero-shot Accuracy ↑} \\ \cmidrule{4-9}
& 
& 
& PQ & ARC-e & ARC-c & HS & WG & Avg. \\ \midrule
 & \texttt{FP16} 
 & -      
 & 79.05 & 74.58 & 46.25 & 76.05 & 68.98 & 68.98
 \\ \cmidrule{2-9}
 & \texttt{W4A4} 
 & Quarot 
 & 76.77 & 69.87 & 40.87 & 72.16 & 63.77 & 64.69
 \\ 
 \cmidrule{2-9}
 7B 
 & \texttt{W4A4} g128
 & Atom 
 & 75.14 & 52.99 & 38.40 & 69.37 & 62.75 & 59.73
 \\ \cmidrule{2-9}
 & \texttt{W4A8KV4} 
 & \cellcolor{mylightblue}\algo 
 & \cellcolor{mylightblue}78.07 & \cellcolor{mylightblue}73.11 & \cellcolor{mylightblue}45.05 & \cellcolor{mylightblue}74.12 & \cellcolor{mylightblue}67.48 & \cellcolor{mylightblue}67.57
 \\ 
 & \makecell[l]{\texttt{W4A8KV4} g128}
 & \cellcolor{mylightblue}\algo
 & \cellcolor{mylightblue}\textbf{78.07} & \cellcolor{mylightblue}\textbf{73.32} & \cellcolor{mylightblue}\textbf{44.80} & \cellcolor{mylightblue}\textbf{74.98} & \cellcolor{mylightblue}\textbf{68.59} & \cellcolor{mylightblue}\textbf{67.95}
 \\ \midrule
 & \texttt{FP16} 
 & -      
 & 80.52 & 77.44 & 49.06 & 79.38 & 72.22 & 71.72
 \\ \cmidrule{2-9}
 & \texttt{W4A4} 
 & Quarot 
 & 78.89 & 72.98 & 46.59 & 76.37 & 70.24 & 69.01
 \\
 \cmidrule{2-9}
 13B 
 & \texttt{W4A4} g128
 & Atom 
 & 76.50 & 57.49 & 42.32 & 73.84 & 67.40 & 63.51
 \\ \cmidrule{2-9}
 & \texttt{W4A8KV4} 
 & \cellcolor{mylightblue}\algo
 & \cellcolor{mylightblue}\textbf{79.71}  & \cellcolor{mylightblue}75.97 & \cellcolor{mylightblue}48.38 & \cellcolor{mylightblue}77.80 & \cellcolor{mylightblue}\textbf{70.96} & \cellcolor{mylightblue}70.56
 \\ 
 & \makecell[l]{\texttt{W4A8KV4} g128}
 & \cellcolor{mylightblue}\algo
 & \cellcolor{mylightblue}79.43  & \cellcolor{mylightblue}\textbf{77.06} & \cellcolor{mylightblue}\textbf{48.81} & \cellcolor{mylightblue}\textbf{78.35} & \cellcolor{mylightblue}70.48 & \cellcolor{mylightblue}\textbf{70.83}
 \\\midrule
 & \texttt{FP16} 
 & -      
 & 82.70 & 81.02 & 57.34 & 83.82 & 77.98 & 76.57
 \\ \cmidrule{2-9}
 & \texttt{W4A4} 
 & Quarot 
 & 82.43 & 80.43 & 56.23 & 81.82 & 76.24 & 75.43
 \\
 \cmidrule{2-9}
 70B
 & \texttt{W4A4} g128
 & Atom 
 & 79.92 & 58.25 & 46.08 & 79.06 & 74.27 & 67.52
 \\ \cmidrule{2-9}
 & \texttt{W4A8KV4} 
 & \cellcolor{mylightblue}\algo
 & \cellcolor{mylightblue}82.64  & \cellcolor{mylightblue}79.80 & \cellcolor{mylightblue}\textbf{56.83}  & \cellcolor{mylightblue}82.78  & \cellcolor{mylightblue}77.51  & \cellcolor{mylightblue}75.91 
 \\ 
 & \makecell[l]{\texttt{W4A8KV4} g128}
 & \cellcolor{mylightblue}\algo
 & \cellcolor{mylightblue}\textbf{82.92}  & \cellcolor{mylightblue}\textbf{80.93} & \cellcolor{mylightblue}56.40 & \cellcolor{mylightblue}\textbf{83.28} & \cellcolor{mylightblue}\textbf{78.45} & \cellcolor{mylightblue}\textbf{76.40}
 \\
 \bottomrule
 \multicolumn{9}{l}{* For reference, using MX-FP4 for \texttt{W4A4} quantizing Llama-7B model will decrease the }\\
 \multicolumn{9}{l}{  accuracy from 72.9 to 63.7 on ARC easy and from 44.7 to 35.5 on ARC challenge task.~\cite{rouhani2023microscaling}}
\end{tabular}
}
\label{tab:evaluation:zero-shot}
\end{table*}

\begin{figure*}[t]
    \centering
    \includegraphics[width=\linewidth]{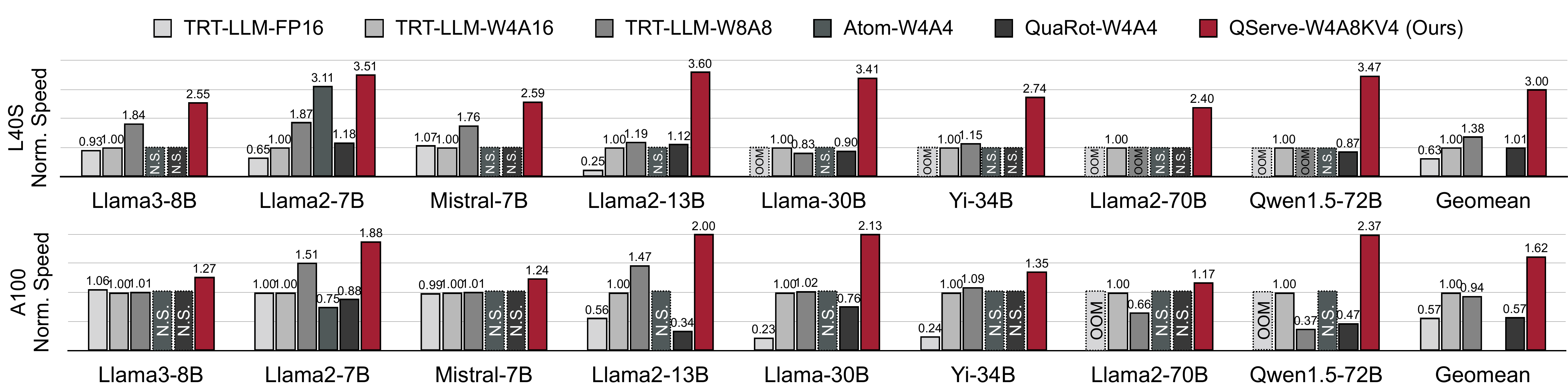}
    \caption{\system significantly outperforms existing large language model (LLM) serving frameworks in batched generation tasks across different LLMs, ranging from 7B to 72B models. It achieves an average speedup of 2.36$\times$ over the state-of-the-art LLM serving system, TensorRT-LLM v0.9.0, on the L40S GPU, and it is also 1.68$\times$ faster on the  A100 GPU. All experiments were conducted under the same device memory budget (\ie 80GB on A100 and 48GB on L40S). We omit the geometric mean speedup of Atom since it only supports Llama2-7B. For absolute values, see Table \ref{tab:abs_main_speed}.}%
    \label{fig:evaluation:main_speed}
\end{figure*}

\begin{table*}[ht]
\centering
\small
\caption{The absolute token generation throughput of \system and TensorRT-LLM in Fig. \ref{fig:evaluation:main_speed}. \textsuperscript{*}: we calculate the speedup over highest achieveable throughput from TensorRT-LLM across all three precision configurations. Our \system system achieves competitive throughput on L40S GPU compared to TensorRT-LLM on A100, effectively reducing the dollar cost of LLM serving by 3$\times$. Unit: tokens/second.}
\scalebox{0.98}{\begin{tabular}{cccccccccc}
\toprule
\multirow{2}{*}{Device} & \multirow{2}{*}{System} & Llama-3 & Llama-2 & Mistral & LLama-2 & LLaMA & Yi & Llama-2 & Qwen1.5 \\
 & & 8B & 7B & 7B & 13B & 30B & 34B & 70B & 72B \\
\midrule

& TRT-LLM-\texttt{FP16} & 1326 & 444 & 1566 & 92 & OOM & OOM & OOM & OOM
\\ \cmidrule{2-10}
& TRT-LLM-\texttt{W4A16} & 1431 & 681 & 1457 & 368 & 148 & 313 & 119 & 17

\\ \cmidrule{2-10}
L40S & TRT-LLM-\texttt{W8A8} & 2634 & 1271 & 2569 & 440 & 123 & 364 & OOM & OOM
\\ \cmidrule{2-10}
 & QServe (Ours) 
 & \textbf{3656} & \textbf{2394} & \textbf{3774} & \textbf{1327} & \textbf{504} & \textbf{869} & \textbf{286} & \textbf{59} 
  \\ \cmidrule{2-10}
 & Speedup\textsuperscript{*} & \textbf{1.39$\times$} & \textbf{1.88$\times$} & \textbf{1.47}$\times$ & \textbf{3.02}$\times$ & \textbf{3.41}$\times$ & \textbf{2.39}$\times$ & \textbf{2.40}$\times$ & \textbf{3.47}$\times$ 
 \\\midrule

& TRT-LLM-\texttt{FP16} & 2503 & 1549 & 2371 & 488 & 80 & 145 & OOM & OOM 
\\ \cmidrule{2-10}

& TRT-LLM-\texttt{W4A16} & 2370 & 1549 & 2403 & 871 & 352 & 569 & 358 & 143 

\\ \cmidrule{2-10}
A100 & TRT-LLM-\texttt{W8A8} & 2396 & 2334 & 2427 & 1277 & 361 & 649 & 234 & 53
 \\ \cmidrule{2-10}
 & QServe (Ours) 
 & \textbf{3005} & \textbf{2908} & \textbf{2970} & \textbf{1741} & \textbf{749} & \textbf{797} & \textbf{419} & \textbf{340}
 \\
 \cmidrule{2-10}
 & Speedup\textsuperscript{*} & \textbf{1.20}$\times$ & \textbf{1.25}$\times$ & \textbf{1.22}$\times$ & \textbf{1.36}$\times$ & \textbf{2.07}$\times$ & \textbf{1.23}$\times$ & \textbf{1.17}$\times$ & \textbf{2.38}$\times$ 
 \\

 \bottomrule
\end{tabular}
}
\label{tab:abs_main_speed}
\vspace{-10pt}
\end{table*}

\vspace{-5pt}
\section{Evaluation}
\label{sect:evaluation}

\subsection{Evaluation Setup}

\textbf{Algorithm.} The \algo quantization algorithm is implemented using HuggingFace~\cite{transformers} on top of PyTorch~\cite{pytorch}. We use per-token symmetric INT8 quantization on activations, and per-head asymmetric INT4 quantization on KV cache. 
``\textbf{\texttt{W4A8KV4}} g128" refers to the case where \system used progressive group quantization on weights: per-channel symmetric INT8 quantization followed by asymmetric INT4 quantization with a group size of 128, while ``\textbf{\texttt{W4A8KV4}}" is the per-channel counterpart for weight quantization. %

\textbf{System.} \system serving system is implemented using CUDA and PTX assembly for high-performance GPU kernels. We also provide a purely PyTorch-based front-end framework for better flexibility. 
For the throughput benchmarking, we perform all experiments under PyTorch 2.2.0 with CUDA 12.2, unless otherwise specified. The throughput numbers reported are real measurements on NVIDIA GPUs. For baseline systems, we use TensorRT-LLM v0.9.0 and latest main branches from QuaRot and Atom as of April 18\textsuperscript{th}, 2024. Paged attention is enabled for all systems except QuaRot, which does not offer corresponding support.

\subsection{Accuracy Evaluation}

\paragraph{Benchmarks.}
We evaluated \algo on the Llama-1~\cite{touvron2023llama}, Llama-2~\cite{touvron2023llama2}, Llama-3 families, Mistral-7B~\cite{jiang2023mistral}, Mixtral-8x7B~\cite{jiang2024mixtral} and Yi-34B~\cite{ai2024yi} models. Following previous literature~\cite{llmint8,gptq,zhao2023atom,ashkboos2024quarot, xiao2023smoothquant,lin2023awq}, we evaluated \algo-quantized models on language modeling and zero-shot tasks. Specifically, we evaluated on WikiText2~\cite{wikitext} for perplexity, and evaluated on PIQA~\cite{piqa} (PQ), ARC~\cite{arc}, HellaSwag~\cite{hellaswag} (HS) and WinoGrande~\cite{sakaguchi2019winogrande} (WG) with \texttt{lm\_eval}~\cite{eval-harness}.

\paragraph{Baselines.}
We compared \algo to widely used post-training LLM quantization techiniques, SmoothQuant~\cite{xiao2023smoothquant}, GPTQ~\cite{gptq}, AWQ~\cite{lin2023awq}, and recently released state-of-the-art 4-bit weight-activation quantization frameworks, Atom~\cite{zhao2023atom} and QuaRot~\cite{ashkboos2024quarot}. For SmoothQuant, we uses static per-tensor symmetric 8-bit quantization for KV cache following the settings in the TensorRT-LLM~\cite{trtllm}.
For GPTQ, we use their latest version with ``reorder" trick, denoted as ``GPTQ-R". For QuaRot and Atom, we mainly evaluated using Pile validation dataset as calibration dataset. We also report their results with WikiText2 as calibration dataset in gray color. For ``\textbf{\texttt{W4A8KV4}} g128" setting, both QuaRot and Atom does not support progressive group quantization, and thus we evaluated them using ordinary group weight quantization (\ie, each group has one FP16 scale factor). Unsupported models and quantization settings will be reported as NaN.

\paragraph{WikiText2 perplexity.}\tab{tab:evaluation:wikitext} compares the Wikitext2 perplexity results between \algo and other baselines. For Llama-2-7B, compared to \textbf{\texttt{W8A8}} SmoothQuant and \textbf{\texttt{W4A16}} AWQ, \algo only increased perplexity by up to 0.16 %
\algo consistently outperformed Atom with either \textbf{\texttt{W4A4}} or \textbf{\texttt{W4A8KV4}} quantization precision. \algo also showed up to 0.49 perplexity improvement compared to \textbf{\texttt{W4A4}} Quarot.

\paragraph{Zero-shot Accuracy and Long-Context Accuracy.}We report the zero-shot accuracy of five common sense tasks in \tab{tab:evaluation:zero-shot}. \algo significantly outperformed other 4-bit quantization methods. 
Especially on the Winogrande task, compared to Quarot, \algo accuracy is 4.82\% higher.
Compared to \texttt{FP16}, \algo only introduced 1.03\%, 0.89\% and 0.40\% accuracy loss for Llama-2 at 7B, 13B and 70B size.
Furthermore, our results in Table~\ref{tab:evaluation:long-context} demonstrate that QoQ can maintain minimal degradation on the long-context performance relative to the \texttt{BF16} baseline.

\begin{table*}[t]
\setlength{\tabcolsep}{0pt}
\centering
\caption{LongBench evaluation for Llama-3.1-8b-Instruct using QoQ W4A8KV4 g128.}
\scalebox{0.9}{
\begin{tabular}{lccccccccccc}
\toprule
 & \rotatebox{25}{DuReader} & \rotatebox{25}{GovReport} & \rotatebox{25}{HotpotQA} & \rotatebox{25}{MultiNews} & \rotatebox{25}{Musique} & \rotatebox{25}{QMSum} & \rotatebox{25}{SamSum} & \rotatebox{25}{TriviaQA} & \rotatebox{25}{TREC}  & \rotatebox{25}{MultiFieldQA-En} & \rotatebox{25}{Average} \\ \midrule
BF16 & 35.07    & 34.54     & 16.68    & 26.84     & 11.68   & 23.48 & 43.50  & 91.65    & 72.50 & 29.22           & 38.52   \\
\cellcolor{mylightblue}QoQ  & \cellcolor{mylightblue}35.45    & \cellcolor{mylightblue}34.09     & \cellcolor{mylightblue}17.46    & \cellcolor{mylightblue}26.73     & \cellcolor{mylightblue}12.05   & \cellcolor{mylightblue}23.45 & \cellcolor{mylightblue}44.42  & \cellcolor{mylightblue}91.45    & \cellcolor{mylightblue}71.00 & \cellcolor{mylightblue}27.65           & \cellcolor{mylightblue}38.38  \\
\bottomrule
\end{tabular}
}
\label{tab:evaluation:long-context}
\vspace{-10pt}
\end{table*}

\begin{figure*}[htbp]
    \centering
    \includegraphics[width=0.95\linewidth]{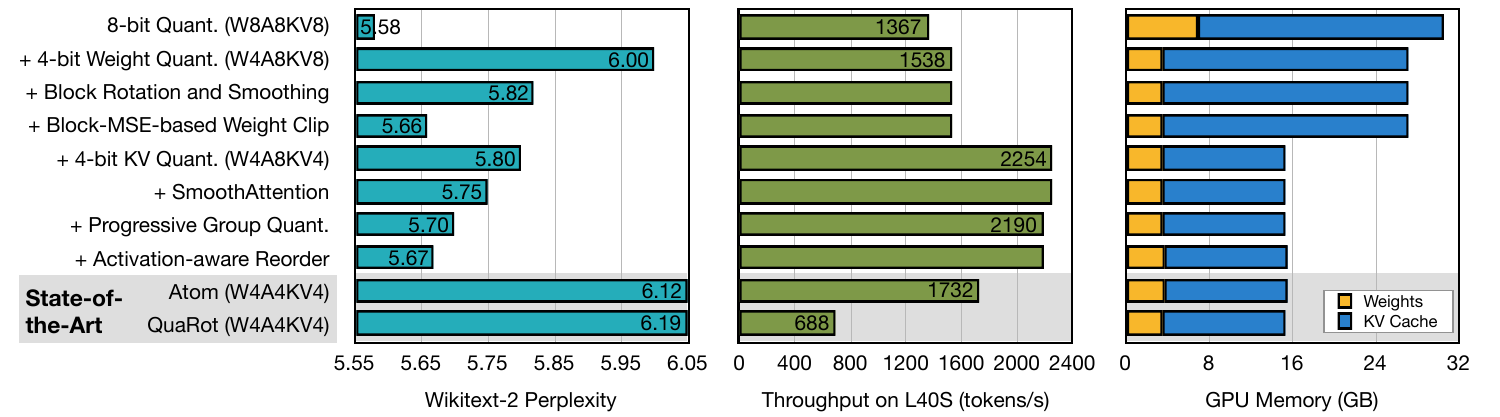}
    \vspace{-10pt}
    \caption{Ablation study on quantization techniques used in \algo and the impact of serving throughput, GPU memory consumption in \system. The model used here is Llama-2-7B.}
    \label{fig:evaluation:alg-ablation}
    \vspace{-10pt}
\end{figure*}

\begin{figure*}[t]
    \vspace{5pt}
    \centering
    \includegraphics[width=\linewidth]{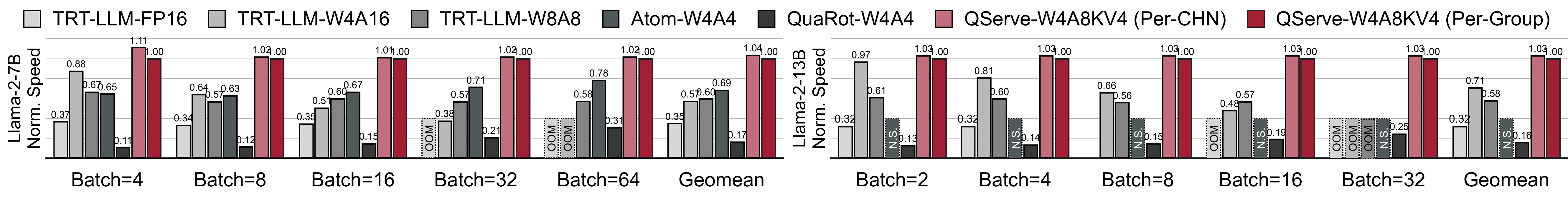}
    \vspace{-25pt}
    \caption{Same-batch throughput comparison between \system and baseline systems on L40S. We use an input sequence length of 1024 and output sequence length of 512.} %
    \label{fig:evaluation:main_speed_same_batch}
    \vspace{-5pt}
\end{figure*}

\subsection{Efficiency Evaluation}

We assessed the efficiency of \system on A100-80G-SXM4 and L40S-48G GPUs by comparing it against TensorRT-LLM (using \texttt{FP16}, \textbf{\texttt{W8A8}}, and \textbf{\texttt{W4A16}} precisions), Atom (\textbf{\texttt{W4A4}}), and QuaRot (\textbf{\texttt{W4A4}}). The primary metric for system evaluation is the maximum achievable throughput within the same memory constraints, where we use an input sequence length of 1024 and output sequence length of 512. We notice that Atom only supports Llama-2-7B, and QuaRot does not support GQA. Therefore, we skip these unsupported models when measuring the performance of baseline systems. 

We present relative performance comparisons in Figure \ref{fig:evaluation:main_speed} and absolute throughput values in Table \ref{tab:abs_main_speed}. We use per-channel quantization for A100 and per-group quantization for L40S. This is because L40S has stronger CUDA cores for dequantization. Relative to the best-performing configuration of TensorRT-LLM, \system demonstrates significant improvements on A100: it achieves \textbf{2\x} higher throughput for Llama-1-30B, \textbf{1.2-1.4\x} higher throughput for Llama-2 models, \textbf{1.2\x} higher throughput for Mistral and Yi, and \textbf{2.4\x} higher throughput for Qwen-1.5. The performance improvements are particularly notable on the L40S GPUs, where we observed a throughput improvement ranging from \textbf{1.47\x} to \textbf{3.47\x} across all seven models evaluated. Remarkably, despite the L40S's significantly smaller memory capacity compared to the A100, \system effectively maintains the same batch size as TensorRT-LLM on the A100. This achievement is attributed to our aggressive 4-bit quantization applied to both weights and the KV cache. By examining Table \ref{tab:abs_main_speed}, we clearly observe that serving five of seven models under 34B on L40S with \system achieves even higher throughput than serving them on A100 using TensorRT-LLM. Our performance gain over Atom and QuaRot on A100 is even more prominent since these systems did not outperform TensorRT-LLM. On L40S, \system still achieves 10\% higher throughput than Atom when running Llama-2-7B, the only model supported by their system despite the fact that we use higher quantization precision. Besides, the accuracy achieved by \system is much better than Atom, as indicated in Table \ref{tab:evaluation:zero-shot}.

\subsection{Analysis and Discussion.}

\textbf{Ablation study on quantization techniques.} We examine the impact on accuracy of various quantization techniques implemented in \algo. 
Our analysis begins with round-to-nearest (RTN) \textbf{\texttt{W8A8}} quantization on Llama-2-7B (per-channel + per-token). We then lower the quantization precision and apply different techniques step-by-step. For each step, we evaluated the WikiText2 perplexity and end-to-end inference performance on L40S with 64 requests of 1024 input tokens and 512 output tokens.
The results are detailed in \fig{fig:evaluation:alg-ablation}.
We see that reducing the weight precision to 4 bits significantly impaired the model performance, though it increased end-to-end processing speed by 1.12\x and saved 3.5GB GPU memory. Rotating the block input modules helped suppress the activation outliers, resulting in 0.18 perplexity improvement. In addition, minimizing the block output MSE through weight clipping further decreased the perplexity by 0.16. Consequently, our \textbf{\texttt{W4A8}} configuration has achieved a perplexity comparable to that of \textbf{\texttt{W4A16}}. However, quantizing KV cache to 4 bits again deteriorated model performance by 0.14, although it substantially enhanced the end-to-end inference throughput by 1.47\x and halved GPU memory usage. \smoothattention reduced perplexity by 0.05, without adding system overhead. Progressive group quantization further improved perplexity by an additional 0.04, with only a negligible increase in dequantization overhead. Lastly, activation-aware channel reordering enhanced perplexity by 0.03.

\begin{figure}[t]
    \centering
    \includegraphics[width=\linewidth]{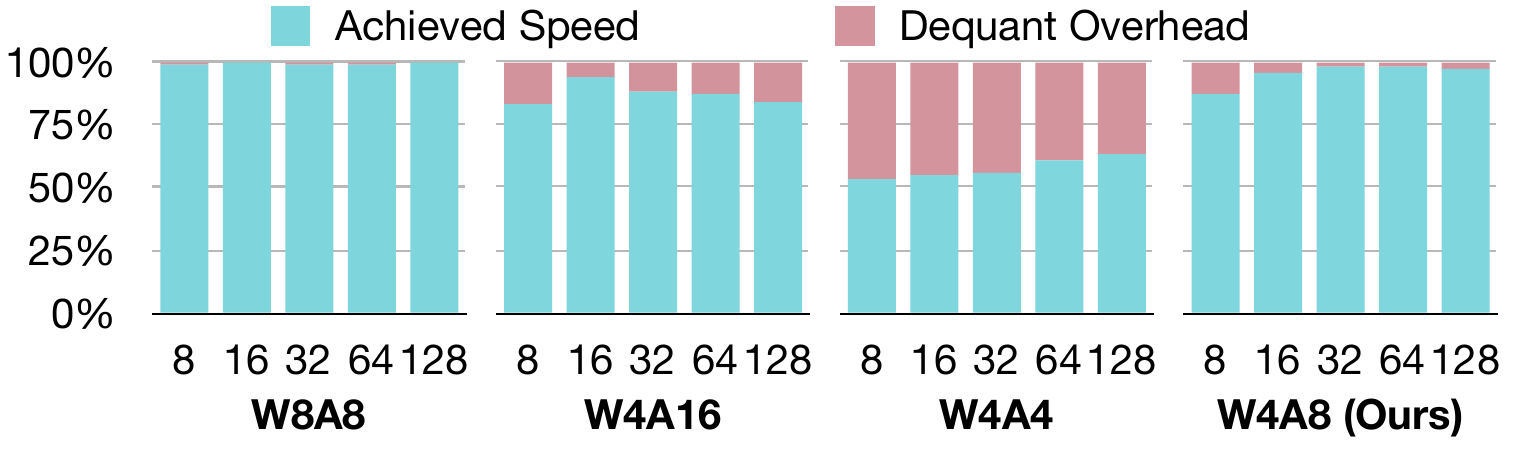}
    \vspace{-25pt}
    \caption{The dequantization overhead in \system is much smaller than that in Atom-\textbf{\texttt{W4A4}} (up to 90\%).}
    \label{fig:dequant_overhead}
    \vspace{-20pt}
\end{figure}

\textbf{Ablation study on \system system: Dequantization overhead.} We measure the dequantization overhead of per-group \system-\textbf{\texttt{W4A8}} GEMM and other baselines in Figure \ref{fig:dequant_overhead}. Our dequantization overhead is comparable with TRT-LLM-\textbf{\texttt{W4A16}}, but since we perform computation on \texttt{INT8} tensor cores, we enjoy 2\x higher throughput. 

\textbf{Comparisons under the same batches.} We demonstrate speedup results under the same batch sizes in Figure \ref{fig:evaluation:main_speed_same_batch}. For Llama-2-7B, we show that the 1.88\x speedup over TRT-LLM can be broken down to two parts: 1.45\x from same batch speedup and 1.3\x from the enlarged batch size. For larger models like Llama-2-13B, scaling up the batch size and single batch speedup are equally important (1.7\x improvement).

\textbf{Improvement breakdown for \texttt{KV4} attention.} We detail the enhancements from attention optimizations in Section \sect{sect:system:attention}. Starting with the basic \texttt{KV4} implementation, which exhibits an A100 latency of 0.48ms for a 64$\times$1024 input, the application of bit tricks from \cite{kim2022says} reduces the kernel latency to 0.44ms. Further improvements are achieved by simplifying the control flow, which reduces latency by an additional 0.05ms. Subsequently, converting the QK and SV products to \texttt{FP16} each contributes a 0.03ms latency reduction. Asynchronous prefetching of dequantization parameters at the start of the attention kernel further enhances performance, ultimately reducing the latency to 0.28ms and achieving an end-to-end improvement of 1.7\x.

\section{Related Work}

\vspace{3pt}
\textbf{Quantization of LLMs.}
Quantization reduces the size of LLMs and speedup inference. There are two primary quantization strategies:
(1) \textbf{Weight-only quantization}~\cite{gptq,lin2023awq,dettmers2023spqr,kim2024squeezellm} benefits edge devices where the workload is memory-bound, improving weight-loading speed. However, for cloud services with high user traffic and required batch processing, this method falls short as it does not accelerate computation in compute-bound scenarios.
(2) \textbf{Weight-activation quantization} accelerates computation in batch processing by quantizing both weights and activations~\cite{llmint8,wei2022outlier,xiao2023smoothquant}. OmniQuant~\cite{OmniQuant} and Atom~\cite{zhao2023atom} exploring more aggressive quantizations (\textbf{\texttt{W4A4, W4A8}}) and mixed precision to enhance model quality and efficiency, though these can impact model accuracy and reduce serving throughput. QuaRot~\cite{ashkboos2024quarot} further refines \textbf{\texttt{W4A4}} by rotating weights and activations at the cost of increased computational overhead due to additional transformations required during inference.

\vspace{3pt}
\textbf{LLM serving systems.}
Numerous systems have been proposed for efficient LLM deployment. Orca~\cite{orca} employs iteration-level scheduling and selective batching in distributed systems. vLLM~\cite{vllm} features virtual memory-inspired PagedAttention, optimizing KV cache management. SGLang~\cite{sglang} enhances LLM programming with advanced primitives and RadixAttention. LMDeploy~\cite{lmdeploy} offers persistent batching and blocked KV cache features to improve deployment efficiency. LightLLM~\cite{lightllm} manages GPU memory with token-wise KV cache control via Token Attention, increasing throughput. MLC-LLM~\cite{mlcllm} utilizes compiler acceleration for versatile LLM deployment across edge devices. TensorRT-LLM~\cite{trtllm} is the leading industry solution and the most important baseline in this paper.

\textbf{LLM Accelerators.} Transformers and LLMs have also generated considerable research interest in domain-specific accelerator design. Several works, such as $A^3$~\cite{ham20203}, ELSA~\cite{ham2021elsa}, and SpAtten~\cite{wang2021spatten}, have applied pruning techniques to the attention module, while GOBO~\cite{zadeh2020gobo} and EdgeBERT~\cite{tambe2021edgebert} have investigated quantization approaches. Additionally, DOTA~\cite{qu2022dota} introduces a lightweight, runtime detector for omitting weak attention connections, coupled with specialized accelerators for transformer inference. Apart from attention optimizations, STA~\cite{fang2022sta} leverages $N$:$M$ sparsity and specialized softmax module to reduce off-chip communication. Moreover, DFX~\cite{hong2022dfx} exploits model parallelism and optimized dataflow for low-latency generation. However, these accelerators have yet to be scaled up to recent LLMs with billions of parameters.

\section{Conclusion}

We introduce \system, an algorithm and system co-design framework tailored to quantize large language models (LLMs) to \textbf{\texttt{W4A8KV4}} precision, facilitating their efficient deployment on GPUs. On the algorithmic front, we design the QoQ quantization method that features progressive quantization, enabling \textbf{\texttt{W4A8}} GEMM operations to be executed on \texttt{INT8} tensor cores, and SmoothAttention, which significantly reduces accuracy loss resulting from \textbf{\texttt{KV4}} quantization. Correspondingly, in the \system system, we leverage the protective range established in the first level of progressive quantization to enable \texttt{INT4} to \texttt{INT8} dequantization. This process utilizes full register-level parallelism and employs a subtraction-after-multiplication computation sequence. Additionally, we implement compute-aware weight reordering to minimize the overhead associated with pointer arithmetic. As a result, when serving seven representative LLMs on A100 and L40S GPUs, \system achieves up to \textbf{2.4-3.5\x} higher throughput over the industrial standard for LLM serving, TensorRT-LLM.

\vspace{-5pt}
\section*{Acknowledgements}
We thank MIT-IBM Watson AI Lab, MIT AI Hardware Program, MIT Amazon Science Hub, and NSF for supporting this research. We also thank Julien Demouth, June Yang, and Dongxu Yang from NVIDIA for their helpful discussions.

\nocite{langley00}

\bibliography{refs}

\begin{thebibliography}{45}
\providecommand{\natexlab}[1]{#1}
\providecommand{\url}[1]{\texttt{#1}}
\expandafter\ifx\csname urlstyle\endcsname\relax
  \providecommand{\doi}[1]{doi: #1}\else
  \providecommand{\doi}{doi: \begingroup \urlstyle{rm}\Url}\fi

\bibitem[Ainslie et~al.(2023)Ainslie, Lee-Thorp, de~Jong, Zemlyanskiy, Lebr{\'o}n, and Sanghai]{ainslie2023gqa}
Ainslie, J., Lee-Thorp, J., de~Jong, M., Zemlyanskiy, Y., Lebr{\'o}n, F., and Sanghai, S.
\newblock Gqa: Training generalized multi-query transformer models from multi-head checkpoints.
\newblock \emph{arXiv preprint arXiv:2305.13245}, 2023.

\bibitem[Ashkboos et~al.(2024)Ashkboos, Mohtashami, Croci, Li, Jaggi, Alistarh, Hoefler, and Hensman]{ashkboos2024quarot}
Ashkboos, S., Mohtashami, A., Croci, M.~L., Li, B., Jaggi, M., Alistarh, D., Hoefler, T., and Hensman, J.
\newblock Quarot: Outlier-free 4-bit inference in rotated llms.
\newblock \emph{arXiv preprint arXiv:2404.00456}, 2024.

\bibitem[Bisk et~al.(2020)Bisk, Zellers, Bras, Gao, and Choi]{piqa}
Bisk, Y., Zellers, R., Bras, R.~L., Gao, J., and Choi, Y.
\newblock Piqa: Reasoning about physical commonsense in natural language.
\newblock In \emph{Thirty-Fourth AAAI Conference on Artificial Intelligence}, 2020.

\bibitem[Chee et~al.(2024)Chee, Cai, Kuleshov, and Sa]{quip}
Chee, J., Cai, Y., Kuleshov, V., and Sa, C.~D.
\newblock Quip: 2-bit quantization of large language models with guarantees, 2024.

\bibitem[Clark et~al.(2018)Clark, Cowhey, Etzioni, Khot, Sabharwal, Schoenick, and Tafjord]{arc}
Clark, P., Cowhey, I., Etzioni, O., Khot, T., Sabharwal, A., Schoenick, C., and Tafjord, O.
\newblock Think you have solved question answering? try arc, the ai2 reasoning challenge, 2018.

\bibitem[Contributors(2023{\natexlab{a}})]{lightllm}
Contributors, L.
\newblock Lightllm: A light and fast inference service for llm.
\newblock \url{https://github.com/ModelTC/lightllm}, 2023{\natexlab{a}}.

\bibitem[Contributors(2023{\natexlab{b}})]{lmdeploy}
Contributors, L.
\newblock Lmdeploy: A toolkit for compressing, deploying, and serving llm.
\newblock \url{https://github.com/InternLM/lmdeploy}, 2023{\natexlab{b}}.

\bibitem[Dettmers et~al.(2022)Dettmers, Lewis, Belkada, and Zettlemoyer]{llmint8}
Dettmers, T., Lewis, M., Belkada, Y., and Zettlemoyer, L.
\newblock {GPT}3.int8(): 8-bit matrix multiplication for transformers at scale.
\newblock In Oh, A.~H., Agarwal, A., Belgrave, D., and Cho, K. (eds.), \emph{Advances in Neural Information Processing Systems}, 2022.

\bibitem[Dettmers et~al.(2023{\natexlab{a}})Dettmers, Pagnoni, Holtzman, and Zettlemoyer]{dettmers2023qlora}
Dettmers, T., Pagnoni, A., Holtzman, A., and Zettlemoyer, L.
\newblock Qlora: Efficient finetuning of quantized llms.
\newblock \emph{arXiv preprint arXiv:2305.14314}, 2023{\natexlab{a}}.

\bibitem[Dettmers et~al.(2023{\natexlab{b}})Dettmers, Svirschevski, Egiazarian, Kuznedelev, Frantar, Ashkboos, Borzunov, Hoefler, and Alistarh]{dettmers2023spqr}
Dettmers, T., Svirschevski, R., Egiazarian, V., Kuznedelev, D., Frantar, E., Ashkboos, S., Borzunov, A., Hoefler, T., and Alistarh, D.
\newblock Spqr: A sparse-quantized representation for near-lossless llm weight compression, 2023{\natexlab{b}}.

\bibitem[Fang et~al.(2022)Fang, Zhou, and Wang]{fang2022sta}
Fang, C., Zhou, A., and Wang, Z.
\newblock An algorithm--hardware co-optimized framework for accelerating n: M sparse transformers.
\newblock \emph{IEEE Transactions on Very Large Scale Integration (VLSI) Systems}, 30\penalty0 (11):\penalty0 1573--1586, 2022.

\bibitem[Frantar et~al.(2022)Frantar, Ashkboos, Hoefler, and Alistarh]{gptq}
Frantar, E., Ashkboos, S., Hoefler, T., and Alistarh, D.
\newblock {GPTQ}: Accurate post-training compression for generative pretrained transformers.
\newblock \emph{arXiv preprint arXiv:2210.17323}, 2022.

\bibitem[Gao et~al.(2023)Gao, Tow, Abbasi, Biderman, Black, DiPofi, Foster, Golding, Hsu, Le~Noac'h, Li, McDonell, Muennighoff, Ociepa, Phang, Reynolds, Schoelkopf, Skowron, Sutawika, Tang, Thite, Wang, Wang, and Zou]{eval-harness}
Gao, L., Tow, J., Abbasi, B., Biderman, S., Black, S., DiPofi, A., Foster, C., Golding, L., Hsu, J., Le~Noac'h, A., Li, H., McDonell, K., Muennighoff, N., Ociepa, C., Phang, J., Reynolds, L., Schoelkopf, H., Skowron, A., Sutawika, L., Tang, E., Thite, A., Wang, B., Wang, K., and Zou, A.
\newblock A framework for few-shot language model evaluation, 12 2023.
\newblock URL \url{https://zenodo.org/records/10256836}.

\bibitem[Ham et~al.(2020)Ham, Jung, Kim, Oh, Park, Song, Park, Lee, Park, Lee, et~al.]{ham20203}
Ham, T.~J., Jung, S.~J., Kim, S., Oh, Y.~H., Park, Y., Song, Y., Park, J.-H., Lee, S., Park, K., Lee, J.~W., et~al.
\newblock A\^{} 3: Accelerating attention mechanisms in neural networks with approximation.
\newblock In \emph{2020 IEEE International Symposium on High Performance Computer Architecture (HPCA)}, pp.\  328--341. IEEE, 2020.

\bibitem[Ham et~al.(2021)Ham, Lee, Seo, Kim, Choi, Jung, and Lee]{ham2021elsa}
Ham, T.~J., Lee, Y., Seo, S.~H., Kim, S., Choi, H., Jung, S.~J., and Lee, J.~W.
\newblock Elsa: Hardware-software co-design for efficient, lightweight self-attention mechanism in neural networks.
\newblock In \emph{2021 ACM/IEEE 48th Annual International Symposium on Computer Architecture (ISCA)}, pp.\  692--705. IEEE, 2021.

\bibitem[Hong et~al.(2022)Hong, Moon, Kim, Lee, Kim, Lee, and Kim]{hong2022dfx}
Hong, S., Moon, S., Kim, J., Lee, S., Kim, M., Lee, D., and Kim, J.-Y.
\newblock Dfx: A low-latency multi-fpga appliance for accelerating transformer-based text generation.
\newblock In \emph{2022 55th IEEE/ACM International Symposium on Microarchitecture (MICRO)}, pp.\  616--630. IEEE, 2022.

\bibitem[Jiang et~al.(2023)Jiang, Sablayrolles, Mensch, Bamford, Chaplot, Casas, Bressand, Lengyel, Lample, Saulnier, et~al.]{jiang2023mistral}
Jiang, A.~Q., Sablayrolles, A., Mensch, A., Bamford, C., Chaplot, D.~S., Casas, D. d.~l., Bressand, F., Lengyel, G., Lample, G., Saulnier, L., et~al.
\newblock Mistral 7b.
\newblock \emph{arXiv preprint arXiv:2310.06825}, 2023.

\bibitem[Jiang et~al.(2024)Jiang, Sablayrolles, Roux, Mensch, Savary, Bamford, Chaplot, Casas, Hanna, Bressand, et~al.]{jiang2024mixtral}
Jiang, A.~Q., Sablayrolles, A., Roux, A., Mensch, A., Savary, B., Bamford, C., Chaplot, D.~S., Casas, D. d.~l., Hanna, E.~B., Bressand, F., et~al.
\newblock Mixtral of experts.
\newblock \emph{arXiv preprint arXiv:2401.04088}, 2024.

\bibitem[Kim et~al.(2024)Kim, Hooper, Gholami, Dong, Li, Shen, Mahoney, and Keutzer]{kim2024squeezellm}
Kim, S., Hooper, C., Gholami, A., Dong, Z., Li, X., Shen, S., Mahoney, M.~W., and Keutzer, K.
\newblock Squeezellm: Dense-and-sparse quantization, 2024.

\bibitem[Kim et~al.(2022)Kim, Henry, Fahim, and Awadalla]{kim2022says}
Kim, Y.~J., Henry, R., Fahim, R., and Awadalla, H.~H.
\newblock Who says elephants can't run: Bringing large scale moe models into cloud scale production.
\newblock \emph{arXiv preprint arXiv:2211.10017}, 2022.

\bibitem[Kwon et~al.(2023{\natexlab{a}})Kwon, Li, Zhuang, Sheng, Zheng, Yu, Gonzalez, Zhang, and Stoica]{kwon2023efficient}
Kwon, W., Li, Z., Zhuang, S., Sheng, Y., Zheng, L., Yu, C.~H., Gonzalez, J., Zhang, H., and Stoica, I.
\newblock Efficient memory management for large language model serving with pagedattention.
\newblock In \emph{Proceedings of the 29th Symposium on Operating Systems Principles}, pp.\  611--626, 2023{\natexlab{a}}.

\bibitem[Kwon et~al.(2023{\natexlab{b}})Kwon, Li, Zhuang, Sheng, Zheng, Yu, Gonzalez, Zhang, and Stoica]{vllm}
Kwon, W., Li, Z., Zhuang, S., Sheng, Y., Zheng, L., Yu, C.~H., Gonzalez, J.~E., Zhang, H., and Stoica, I.
\newblock Efficient memory management for large language model serving with pagedattention.
\newblock In \emph{Proceedings of the ACM SIGOPS 29th Symposium on Operating Systems Principles}, 2023{\natexlab{b}}.

\bibitem[Lin et~al.(2024)Lin, Tang, Tang, Yang, Chen, Wang, Xiao, Dang, Gan, and Han]{lin2023awq}
Lin, J., Tang, J., Tang, H., Yang, S., Chen, W.-M., Wang, W.-C., Xiao, G., Dang, X., Gan, C., and Han, S.
\newblock Awq: Activation-aware weight quantization for llm compression and acceleration.
\newblock In \emph{MLSys}, 2024.

\bibitem[Merity et~al.(2016)Merity, Xiong, Bradbury, and Socher]{wikitext}
Merity, S., Xiong, C., Bradbury, J., and Socher, R.
\newblock Pointer sentinel mixture models, 2016.

\bibitem[NVIDIA(2023)]{trtllm}
NVIDIA.
\newblock {TensorRT-LLM: A TensorRT Toolbox for Optimized Large Language Model Inference}, 2023.
\newblock URL \url{https://github.com/NVIDIA/TensorRT-LLM}.

\bibitem[Paszke et~al.(2019)Paszke, Gross, Massa, Lerer, Bradbury, Chanan, Killeen, Lin, Gimelshein, Antiga, Desmaison, Köpf, Yang, DeVito, Raison, Tejani, Chilamkurthy, Steiner, Fang, Bai, and Chintala]{pytorch}
Paszke, A., Gross, S., Massa, F., Lerer, A., Bradbury, J., Chanan, G., Killeen, T., Lin, Z., Gimelshein, N., Antiga, L., Desmaison, A., Köpf, A., Yang, E., DeVito, Z., Raison, M., Tejani, A., Chilamkurthy, S., Steiner, B., Fang, L., Bai, J., and Chintala, S.
\newblock Pytorch: An imperative style, high-performance deep learning library, 2019.

\bibitem[Qu et~al.(2022)Qu, Liu, Tu, Chen, Ding, and Xie]{qu2022dota}
Qu, Z., Liu, L., Tu, F., Chen, Z., Ding, Y., and Xie, Y.
\newblock Dota: detect and omit weak attentions for scalable transformer acceleration.
\newblock In \emph{Proceedings of the 27th ACM International Conference on Architectural Support for Programming Languages and Operating Systems}, pp.\  14--26, 2022.

\bibitem[Rouhani et~al.(2023)Rouhani, Zhao, More, Hall, Khodamoradi, Deng, Choudhary, Cornea, Dellinger, Denolf, et~al.]{rouhani2023microscaling}
Rouhani, B.~D., Zhao, R., More, A., Hall, M., Khodamoradi, A., Deng, S., Choudhary, D., Cornea, M., Dellinger, E., Denolf, K., et~al.
\newblock Microscaling data formats for deep learning.
\newblock \emph{arXiv preprint arXiv:2310.10537}, 2023.

\bibitem[Sakaguchi et~al.(2019)Sakaguchi, Bras, Bhagavatula, and Choi]{sakaguchi2019winogrande}
Sakaguchi, K., Bras, R.~L., Bhagavatula, C., and Choi, Y.
\newblock Winogrande: An adversarial winograd schema challenge at scale.
\newblock \emph{arXiv preprint arXiv:1907.10641}, 2019.

\bibitem[Shao et~al.(2023)Shao, Chen, Zhang, Xu, Zhao, Li, Zhang, Gao, Qiao, and Luo]{OmniQuant}
Shao, W., Chen, M., Zhang, Z., Xu, P., Zhao, L., Li, Z., Zhang, K.~Z., Gao, P., Qiao, Y., and Luo, P.
\newblock Omniquant: Omnidirectionally calibrated quantization for large language models.
\newblock \emph{arXiv preprint arXiv:2308.13137}, 2023.

\bibitem[Tambe et~al.(2021)Tambe, Hooper, Pentecost, Jia, Yang, Donato, Sanh, Whatmough, Rush, Brooks, et~al.]{tambe2021edgebert}
Tambe, T., Hooper, C., Pentecost, L., Jia, T., Yang, E.-Y., Donato, M., Sanh, V., Whatmough, P., Rush, A.~M., Brooks, D., et~al.
\newblock Edgebert: Sentence-level energy optimizations for latency-aware multi-task nlp inference.
\newblock In \emph{MICRO-54: 54th Annual IEEE/ACM International Symposium on Microarchitecture}, pp.\  830--844, 2021.

\bibitem[team(2023)]{mlcllm}
team, M.
\newblock {MLC-LLM}, 2023.
\newblock URL \url{https://github.com/mlc-ai/mlc-llm}.

\bibitem[Touvron et~al.(2023{\natexlab{a}})Touvron, Lavril, Izacard, Martinet, Lachaux, Lacroix, Rozière, Goyal, Hambro, Azhar, Rodriguez, Joulin, Grave, and Lample]{touvron2023llama}
Touvron, H., Lavril, T., Izacard, G., Martinet, X., Lachaux, M.-A., Lacroix, T., Rozière, B., Goyal, N., Hambro, E., Azhar, F., Rodriguez, A., Joulin, A., Grave, E., and Lample, G.
\newblock Llama: Open and efficient foundation language models, 2023{\natexlab{a}}.

\bibitem[Touvron et~al.(2023{\natexlab{b}})Touvron, Martin, Stone, Albert, Almahairi, Babaei, Bashlykov, Batra, Bhargava, Bhosale, et~al.]{touvron2023llama2}
Touvron, H., Martin, L., Stone, K., Albert, P., Almahairi, A., Babaei, Y., Bashlykov, N., Batra, S., Bhargava, P., Bhosale, S., et~al.
\newblock Llama 2: Open foundation and fine-tuned chat models.
\newblock \emph{arXiv preprint arXiv:2307.09288}, 2023{\natexlab{b}}.

\bibitem[Wang et~al.(2021)Wang, Zhang, and Han]{wang2021spatten}
Wang, H., Zhang, Z., and Han, S.
\newblock Spatten: Efficient sparse attention architecture with cascade token and head pruning.
\newblock In \emph{2021 IEEE International Symposium on High-Performance Computer Architecture (HPCA)}, pp.\  97--110. IEEE, 2021.

\bibitem[Wei et~al.(2022)Wei, Zhang, Zhang, Gong, Zhang, Zhang, Yu, and Liu]{wei2022outlier}
Wei, X., Zhang, Y., Zhang, X., Gong, R., Zhang, S., Zhang, Q., Yu, F., and Liu, X.
\newblock Outlier suppression: Pushing the limit of low-bit transformer language models.
\newblock \emph{arXiv preprint arXiv:2209.13325}, 2022.

\bibitem[Wolf et~al.(2020)Wolf, Debut, Sanh, Chaumond, Delangue, Moi, Cistac, Rault, Louf, Funtowicz, Davison, Shleifer, von Platen, Ma, Jernite, Plu, Xu, Scao, Gugger, Drame, Lhoest, and Rush]{transformers}
Wolf, T., Debut, L., Sanh, V., Chaumond, J., Delangue, C., Moi, A., Cistac, P., Rault, T., Louf, R., Funtowicz, M., Davison, J., Shleifer, S., von Platen, P., Ma, C., Jernite, Y., Plu, J., Xu, C., Scao, T.~L., Gugger, S., Drame, M., Lhoest, Q., and Rush, A.~M.
\newblock Huggingface's transformers: State-of-the-art natural language processing, 2020.

\bibitem[Xiao et~al.(2023)Xiao, Lin, Seznec, Wu, Demouth, and Han]{xiao2023smoothquant}
Xiao, G., Lin, J., Seznec, M., Wu, H., Demouth, J., and Han, S.
\newblock {S}mooth{Q}uant: Accurate and efficient post-training quantization for large language models.
\newblock In \emph{Proceedings of the 40th International Conference on Machine Learning}, 2023.

\bibitem[Young et~al.(2024)Young, Chen, Li, Huang, Zhang, Zhang, Li, Zhu, Chen, Chang, Yu, Liu, Liu, Yue, Yang, Yang, Yu, Xie, Huang, Hu, Ren, Niu, Nie, Xu, Liu, Wang, Cai, Gu, Liu, and Dai]{ai2024yi}
Young, A., Chen, B., Li, C., Huang, C., Zhang, G., Zhang, G., Li, H., Zhu, J., Chen, J., Chang, J., Yu, K., Liu, P., Liu, Q., Yue, S., Yang, S., Yang, S., Yu, T., Xie, W., Huang, W., Hu, X., Ren, X., Niu, X., Nie, P., Xu, Y., Liu, Y., Wang, Y., Cai, Y., Gu, Z., Liu, Z., and Dai, Z.
\newblock Yi: Open foundation models by 01.ai, 2024.

\bibitem[Yu et~al.(2022)Yu, Jeong, Kim, Kim, and Chun]{orca}
Yu, G.-I., Jeong, J.~S., Kim, G.-W., Kim, S., and Chun, B.-G.
\newblock Orca: A distributed serving system for {Transformer-Based} generative models.
\newblock In \emph{16th USENIX Symposium on Operating Systems Design and Implementation (OSDI 22)}, pp.\  521--538, Carlsbad, CA, July 2022. USENIX Association.
\newblock ISBN 978-1-939133-28-1.
\newblock URL \url{https://www.usenix.org/conference/osdi22/presentation/yu}.

\bibitem[Zadeh et~al.(2020)Zadeh, Edo, Awad, and Moshovos]{zadeh2020gobo}
Zadeh, A.~H., Edo, I., Awad, O.~M., and Moshovos, A.
\newblock Gobo: Quantizing attention-based nlp models for low latency and energy efficient inference.
\newblock In \emph{2020 53rd Annual IEEE/ACM International Symposium on Microarchitecture (MICRO)}, pp.\  811--824. IEEE, 2020.

\bibitem[Zellers et~al.(2019)Zellers, Holtzman, Bisk, Farhadi, and Choi]{hellaswag}
Zellers, R., Holtzman, A., Bisk, Y., Farhadi, A., and Choi, Y.
\newblock Hellaswag: Can a machine really finish your sentence?
\newblock \emph{CoRR}, abs/1905.07830, 2019.
\newblock URL \url{http://arxiv.org/abs/1905.07830}.

\bibitem[Zhang et~al.(2023)Zhang, Fei, Wu, He, Lou, and Zhou]{zhang2023dual}
Zhang, L., Fei, W., Wu, W., He, Y., Lou, Z., and Zhou, H.
\newblock Dual grained quantization: Efficient fine-grained quantization for llm.
\newblock \emph{arXiv preprint arXiv:2310.04836}, 2023.

\bibitem[Zhao et~al.(2023)Zhao, Lin, Zhu, Ye, Chen, Zheng, Ceze, Krishnamurthy, Chen, and Kasikci]{zhao2023atom}
Zhao, Y., Lin, C.-Y., Zhu, K., Ye, Z., Chen, L., Zheng, S., Ceze, L., Krishnamurthy, A., Chen, T., and Kasikci, B.
\newblock Atom: Low-bit quantization for efficient and accurate llm serving.
\newblock In \emph{MLSys}, 2023.

\bibitem[Zheng et~al.(2023)Zheng, Yin, Xie, Huang, Sun, Yu, Cao, Kozyrakis, Stoica, Gonzalez, Barrett, and Sheng]{sglang}
Zheng, L., Yin, L., Xie, Z., Huang, J., Sun, C., Yu, C.~H., Cao, S., Kozyrakis, C., Stoica, I., Gonzalez, J.~E., Barrett, C., and Sheng, Y.
\newblock Efficiently programming large language models using sglang, 2023.

\end{thebibliography}
\bibliographystyle{mlsys2024}

\newpage
\appendix

\section{Artifact Appendix}

\subsection{Abstract}

This artifact contains necessary scripts and dependencies
to faithfully reproduce the crucial experiments presented in the paper. To successfully run the experiments, a host system with x86\_64 CPUs is required, along with at least one A100 or L40S NVIDIA GPU. We also provide a pre-built docker image to simplify the environment setup process.

\subsection{Artifact check-list (meta-information)}

{\small
\begin{itemize}
  \item {\bf Program:} Efficiency benchmarking code for QServe; efficiency benchmarking code for baseline systems such as TensorRT-LLM.
  \item {\bf Compilation:} Completed in the docker.
  \item {\bf Transformations:} N/A. 
  \item {\bf Binary:} N/A.
  \item {\bf Model:} Llama-3-8B, Llama-2-7B, Mistral-7B, Llama-2-13B.
  \item {\bf Data set:} None.
  \item {\bf Run-time environment:} NVIDIA Container Toolkit (nvidiadocker).
  \item {\bf Hardware:} A host with x86\_64 CPUs and at least one NVIDIA A100 GPU (recommended) or L40S GPU.
  \item {\bf Run-time state:} N/A.
  \item {\bf Execution:} All benchmarks are executed on NVIDIA GPUs, while some data pre-processing code is executed on the host CPU.
  \item {\bf Metrics:} LLM generation throughput.
  \item {\bf Output:} Generation throughput (tokens/second).
  \item {\bf Experiments:} Inference speed measurement for QServe and baseline systems such as TensorRT-LLM.
  \item {\bf How much disk space required (approximately)?:} 512G.
  \item {\bf How much time is needed to prepare workflow (approximately)?:} Around 1 hour to pull docker images depending on the Internet connection and CPU performance.
  \item {\bf How much time is needed to complete experiments (approximately)?:} Around 1 hour to finish the efficiency benchmarks of QServe; and 2-4 GPU hours to finish the TensorRT-LLM benchmarks depending on the GPU performance and number of tasks to evaluate.
  \item {\bf Publicly available?:} Yes.
  \item {\bf Code licenses (if publicly available)?:} Apache License 2.0.
  \item {\bf Data licenses (if publicly available)?:} MIT.
  \item {\bf Workflow framework used?:} Docker.
  \item {\bf Archived (provide DOI)?:} \url{https://doi.org/10.5281/zenodo.14991385}
\end{itemize}

\subsection{Description}

\vspace{-3pt}
\subsubsection{How delivered}
\vspace{-3pt}

We will provide AE reviewers with a pre-built docker image containing \system, TensorRT-LLM and all necessary dependencies.

\vspace{-5pt}
\subsubsection{Hardware dependencies}
\vspace{-3pt}

A host machine with x86\_64 CPUs and at least one NVIDIA A100 GPU (recommended) or L40S GPU.

\vspace{-5pt}
\subsubsection{Software dependencies}
\vspace{-3pt}

A GPU-compatible Docker runtime environment is required.

\vspace{-5pt}
\subsection{Installation}

We recommend that users utilize our pre-built Docker images to set up the environment and run all experiments within the GPU-supported Docker container.

\begin{lstlisting}[style=mystyle, language=bash]
docker run --gpus all -it --workdir /root shang12138/qserve-mlsys25-ae
\end{lstlisting}

\vspace{-10pt}
\subsection{Experiment workflow}

The generation throughputs of QServe and baseline system (i.e., TensorRT-LLM) can be measured with the following commands.

\begin{lstlisting}[style=mystyle, language=bash]
# QServe benchmark
cd /root/QServe    
bash scripts/benchmark/benchmark_a100.sh 
# Run this command if you run on A100 GPU
# Results in ./parsed_results.csv

# TensorRT-LLM benchmark
cd /root/TensorRT-LLM   
bash launch-all.sh   
# Launch TensorRT-LLM evaluation
# Results in ./results.csv
\end{lstlisting}

\vspace{-10pt}
\subsection{Evaluation and expected result}
\vspace{-15pt}
\begin{table}[h]
\centering
\caption{Generation throughput of QServe and baseline (TensorRT-LLM). Unit: tokens/second.}
\begin{tabular}{cccc}
\toprule
Model       & TensorRT-LLM (W8A8KV8) & QServe     \\ \midrule
Llama-3-8B  & 2387.55  & 2980.69    \\
Llama-2-7B  & 2339.97  & 2860.01    \\
Mistral-7B  & 2427.64  & 3031.93    \\ \bottomrule
\end{tabular}
\label{appendix:expected_results}
\vspace{-5pt}
\end{table}

Table~\ref{appendix:expected_results} provides reference numbers for \system benchmarks. Please note that absolute throughput measurements may vary slightly, even on identical GPU platforms, due to differences in machine conditions. However, the relative acceleration ratios should remain consistent.

\subsection{Experiment customization}

The users are encouraged to carry out experiments with different models and batch sizes by modifying the benchmarking scripts.
Accuracy evaluation is omitted to simplify the environment setup. The accuracy results can be reproduced with open-source library \href{https://github.com/mit-han-lab/deepcompressor/tree/dev/v0.1.0}{\texttt{deepcompressor}}.

\subsection{Methodology}

Submission, reviewing and badging methodology:
\vspace{-10pt}
{\small
\begin{itemize}
  \item \url{http://cTuning.org/ae/submission-20190109.html}
  \vspace{-5pt}
  \item \url{http://cTuning.org/ae/reviewing-20190109.html}
  \vspace{-5pt}
  \item \url{https://www.acm.org/publications/policies/artifact-review-badging}
\end{itemize}
}

\end{document}